\documentclass{article}

     \PassOptionsToPackage{numbers, compress}{natbib}

     \usepackage[final]{neurips_2021}

\usepackage[utf8]{inputenc} 
\usepackage[T1]{fontenc}    
\usepackage{hyperref}       
\usepackage{url}            
\usepackage{booktabs}       
\usepackage{amsfonts}       
\usepackage{nicefrac}       
\usepackage{microtype}      
\usepackage{xcolor}         

\usepackage{makecell}
\usepackage{graphicx}
\usepackage{xspace}
\xspaceaddexceptions{[]\{\}}
\usepackage{algorithm}
\usepackage{algorithmic}
\usepackage{tablefootnote}

\usepackage{colortbl}
\definecolor{bgcolor}{rgb}{1.0, 0.8, 0.5}
\usepackage{amssymb}
\usepackage{amsmath}
\usepackage{mathrsfs}
\usepackage{bm}

\allowdisplaybreaks
\newtheorem{theorem}{Theorem}
\newtheorem{lemma}{Lemma}

\newtheorem{definition}{Definition}
\newtheorem{assumption}{Assumption}

{\hspace*{\fill}$\Box$\par}
{\hspace*{\fill}$\Box$\par\vspace{4mm}}
{\hspace*{\fill}$\Box$\par}

\newcommand*{\qedb}{\hfill\ensuremath{\square}}

\newcommand{\canita}{{\sf CANITA}\xspace}
\newcommand{\algname}[1]{{\sf #1}\xspace}

\newcommand{\eqdef}{:=}
\newcommand{\R}{{\mathbb R}}
\newcommand{\E}{{\mathbb E}}
\newcommand{\Exp}[1]{{\mathbb E}\left[ #1 \right]}

\newcommand{\EB}[1]{\mathbb{E}\left[ #1 \right]}

\newcommand{\ns}[1]{\| #1 \|^2}

\newcommand{\nsB}[1]{\left\| #1 \right\|^2}

\newcommand{\n}[1]{\| #1 \|}
\newcommand{\norm}[1]{\left\| #1 \right\|}
\newcommand{\inner}[2]{\langle #1, #2 \rangle}

\newcommand{\calC}{{\mathcal C}}
\newcommand{\cC}{{\cal C}}

\newcommand{\eat}[1]{}
\newcommand{\red}[1]{\textcolor{red}{#1}}
\newcommand{\blue}[1]{\textcolor{blue}{#1}}
\newcommand{\yes}{\blue{\checkmark}}
\newcommand{\no}{\red{\bm{$\times$~}}}

\newcommand{\hx}{\widehat{x}}

\newcommand{\ux}{y}
\newcommand{\uxt}{y^t}

\newcommand{\xt}{x^t}
\newcommand{\xtn}{x^{t+1}}

\newcommand{\wt}{w^t}
\newcommand{\wtn}{w^{t+1}}

\newcommand{\bx}{z} 

\newcommand{\bxtn}{z^{t+1}}

\newcommand{\allht}{h^t}
\newcommand{\hit}{h_i^t}
\newcommand{\htn}{h^{t+1}}
\newcommand{\hitn}{h_i^{t+1}}

\newcommand{\cit}{\cC_i^{t}}
\newcommand{\gt}{g^{t}}
\newcommand{\etat}{\eta_{t}}
\newcommand{\thetat}{\theta_{t}}
\newcommand{\alphat}{\alpha_{t}}
\newcommand{\pt}{p_{t}}
\newcommand{\betat}{\beta_{t}}
\newcommand{\gammat}{\gamma_{t}}
\newcommand{\tnabla}{g^t}
\newcommand{\cht}{\sum_{i=1}^n \ns{\nabla f_i(\wt) - \hit}}

\newcommand{\etaT}{\eta_{T}}
\newcommand{\thetaT}{\theta_{T}}

\newcommand{\pT}{p_{T}}

\newcommand{\gammaT}{\gamma_{T}}

\title{\canita: Faster Rates for Distributed Convex Optimization with Communication Compression}

\author{%
	Zhize Li\\
	KAUST\\
	\texttt{zhize.li@kaust.edu.sa} \\
	\And
	Peter Richt{\'a}rik \\
	KAUST \\
	\texttt{peter.richtarik@kaust.edu.sa} \\
}

\begin{document}

\maketitle

\begin{abstract}
  Due to the high communication cost in distributed and federated learning, methods relying on compressed communication are becoming increasingly popular. 
  Besides, the best theoretically and practically performing gradient-type methods invariably rely on some form of acceleration/momentum to reduce the number of communications (faster convergence), e.g., Nesterov's accelerated gradient descent~\citep{nesterov1, NesterovBook} and \algname{Adam}~\citep{ADAM}.
  In order to combine the benefits of communication compression and convergence acceleration, we propose a \emph{compressed and accelerated} gradient method based on \algname{ANITA}~\cite{li2021anita} for distributed optimization, which we call \canita.
  Our \canita achieves the \emph{first accelerated rate} {\footnotesize $ O\bigg(\sqrt{\Big(1+\sqrt{\frac{\omega^3}{n}}\Big)\frac{L}{\epsilon}} + \omega\big(\frac{1}{\epsilon}\big)^{\frac{1}{3}}\bigg)$}, which improves upon the state-of-the-art non-accelerated rate  $O\left((1+\frac{\omega}{n})\frac{L}{\epsilon} + \frac{\omega^2+\omega}{\omega+n}\frac{1}{\epsilon}\right)$ of \algname{DIANA}~\citep{khaled2020unified} for distributed general convex problems, where $\epsilon$ is the target error,  $L$ is the smooth parameter of the objective, $n$ is the number of machines/devices, and $\omega$ is the compression parameter (larger $\omega$ means more compression can be applied, and no compression implies $\omega=0$). 
  Our results show that as long as the number of devices $n$ is large (often true in distributed/federated learning), or the compression $\omega$ is not very high,  \canita achieves the faster convergence rate $O\Big(\sqrt{\frac{L}{\epsilon}}\Big)$, i.e., the number of communication rounds is $O\Big(\sqrt{\frac{L}{\epsilon}}\Big)$ (vs.~$O\big(\frac{L}{\epsilon}\big)$ achieved by previous works). 
  As a result, \canita enjoys the advantages of both compression (compressed communication in each round) and acceleration (much fewer communication rounds). 
\end{abstract}

\section{Introduction}
\label{sec:intro}

With the proliferation of edge devices, such as mobile phones, wearables and smart home appliances, comes an increase in the amount of data rich in potential information which can be mined for the benefit of humankind. One of the approaches of turning the raw data into information is via federated learning \citep{FEDLEARN, FL2017-AISTATS}, where typically a single global supervised model is trained in a massively distributed manner over a network of heterogeneous devices.

Training supervised distributed/federated learning models is typically performed by solving an optimization problem of the form
\vspace{-2mm}
\begin{equation}
\vspace{-1.8mm}
\label{eq:prob}
\min_{x\in\R^d} \Big\{f(x):=\frac{1}{n}\sum \limits_{i=1}^{n}f_i(x)\Big\},
\end{equation}
where $n$ denotes the number of devices/machines/workers/clients, and $f_i:\R^d\rightarrow \R$ is a  loss function associated with the data stored on device $i$. We will write $$x^* \eqdef \arg\min_{x\in \R^d} f(x).$$
If more than one minimizer exist,  $x^*$ denotes an arbitrary but fixed solution. We will rely on the solution concept captured in the following definition:
\begin{definition}\label{def:1}
	A random vector $\hx \in \R^d$ is called an $\epsilon$-solution of the distributed problem \eqref{eq:prob} if 
	$$\Exp{f(\hx)}-f(x^*) \leq \epsilon,$$
	where the expectation is with respect to the randomness inherent in the algorithm used to produce $\hx$.
\end{definition}

In distributed and federated learning problems of the form \eqref{eq:prob}, communication of messages across the network typically forms the key bottleneck of the training system.  In the modern  practice of supervised learning in general and deep learning in particular, this is exacerbated by the reliance on massive models described by millions or even billions of parameters. For these reasons, it is very important to devise novel and more efficient training algorithms capable of decreasing the overall communication cost, which can be formalized as the product of the number of communication rounds necessary to train a model of sufficient quality, and the computation and communication cost associated with a typical communication round.

\subsection{Methods with compressed communication}
One of the most common strategies for improving communication complexity is {\em communication compression}~\citep{Seide2015:1bit, alistarh2017qsgd, tonko, Cnat, DIANA, DIANA2, li2020acceleration, li2020unified}.  This strategy is based on the reduction of the size of communicated messages via the application of a suitably chosen lossy compression mechanism, saving precious time spent in each communication round, and hoping that this will not increase the total number of communication rounds. 

Several recent theoretical results suggest that by combining an appropriate (randomized) compression operator with a suitably designed gradient-type method, one can obtain improvement in the total communication complexity over comparable baselines not performing any compression. For instance, this is the case for distributed compressed gradient descent (\algname{CGD})~\citep{alistarh2017qsgd, khirirat2018distributed, Cnat, li2020unified}, and distributed \algname{CGD} methods which employ variance reduction to tame the variance introduced by compression~\citep{SEGA, DIANA, DIANA2, li2020unified, gorbunov2021marina}.

\subsection{Methods with acceleration}

The acceleration/momentum of gradient-type methods is widely-studied in standard optimization problems, which aims to achieve faster convergence rates (fewer communication rounds)~\citep{polyak1964some, nesterov1, NesterovBook, lan2015optimal, lin2015universal, allen2017katyusha, Zhize2019unified, L-SVRG, li2020fast, li2021anita}.
Deep learning practitioners typically rely on  \algname{Adam}~\citep{ADAM}, or one of its many variants, which besides other tricks also adopts momentum.
In particular, \algname{ANITA}~\citep{li2021anita} obtains the current state-of-the-art convergence results for convex optimization. In this paper, we will adopt the acceleration from \algname{ANITA}~\citep{li2021anita} to the distributed setting with compression.

%

\subsection{Can communication compression and acceleration be combined?}

Encouraged by the recent theoretical success of communication compression, and the widespread success of accelerated methods, in this paper we seek to further enhance \algname{CGD} methods with acceleration/momentum, with the aim to obtain provable improvements in overall communication complexity. 

\begin{quote}\emph{Can distributed gradient-type methods theoretically benefit from the combination of gradient compression and acceleration/momentum? To the best of our knowledge, no such results exist in the general convex regime, and in this paper we  close this gap by designing a method that can provably enjoy the advantages of both compression (compressed communication in each round) and acceleration (much fewer communication rounds). }\end{quote}

While there is abundance of research studying communication compression and acceleration in isolation, 
there is very limited work on the combination of both approaches. The first successful combination of gradient compression and acceleration/momentum was recently achieved by the \algname{ADIANA} method of \citet{li2020acceleration}. However, \citet{li2020acceleration} only provide  theoretical results for strongly convex problems, and their method is not applicable to (general) convex problems. So, one needs to both design a new method to handle the convex case, and perform its analysis. A-priori, it is not clear at all what approach would work.

To the best of our knowledge, besides the initial work \citep{li2020acceleration},  we are only aware of two other works for addressing this question \citep{ye2020deed, qian2020error}.  However, both these works still only focus on the simpler and less practically relevant {\em strongly convex} setting.  Thus, this line of research is still largely unexplored. For instance, the well-known logistic regression problem is convex but not strongly convex. Finally, even if a problem is strongly convex, the modulus of strong convexity  is typically not known, or hard to estimate properly.

\begin{table}[t]
	\centering
	\caption{\small Convergence rates for finding an $\epsilon$-solution $\E[f(x^T)]-f(x^*)\leq \epsilon$ of distributed problem \eqref{eq:prob}}
	\label{table:1}
	\small
	\renewcommand{\arraystretch}{2}
	\vspace{1mm}
	\begin{tabular}{|c|c|c|c|}
		\hline \bf
		Algorithms & \bf Strongly convex \tablefootnote{In this strongly convex column, $\kappa:=\frac{L}{\mu}$ denotes the condition number, where $L$ is the smooth parameter and $\mu>0$ is the strong convexity parameter.}   &\bf General convex & \bf Remark\\
		\hline
		\makecell{\algname{QSGD} \citep{alistarh2017qsgd}}
		& ---
		&  $O\left(\frac{L}{\epsilon} + \frac{\omega G^2}{n}\frac{1}{\epsilon^2}\right)$ \tablefootnote{Here \algname{\algname{QSGD}} \citep{alistarh2017qsgd} needs an additional bounded gradient assumption, i.e., $\ns{\nabla f_i(x)}\leq G^2$, $\forall i \in [n],x\in \R^d$.} 
		& \makecell{\yes compression \\ 
			\no acceleration } \\ \hline
		
		\makecell{\hspace{-1mm}\algname{DIANA} \citep{DIANA}\hspace{-4mm}}
		& $O\left(\Big(\big(1+\frac{\omega}{n}\big)\kappa + \omega\Big)\log \frac{1}{\epsilon}\right)$ 
		& --- 
		& \makecell{\yes compression \\ 
			\no acceleration } \\ \hline
		
		\makecell{\algname{DIANA} \citep{DIANA2}}
		& $O\left(\Big(\big(1+\frac{\omega}{n}\big)\kappa + \omega\Big)\log \frac{1}{\epsilon}\right)$ 
		& $O\left(\big(1+\frac{\omega}{n}\big)\frac{L}{\epsilon} + \frac{\omega}{\epsilon}\right)$
		& \makecell{\yes compression \\ 
			\no acceleration } \\ \hline
		
		\makecell{\algname{DIANA} \citep{khaled2020unified}}
		& ---
		& $O\left(\big(1+\frac{\omega}{n}\big)\frac{L}{\epsilon} + \frac{\omega^2 + \omega}{\omega+n}\frac{1}{\epsilon}\right)$
		& \makecell{\yes compression \\ 
			\no acceleration } \\ \hline
		
		\makecell{\algname{ADIANA} \citep{li2020acceleration}}
		& \hspace{-2mm} $O${\scriptsize $\left(\Big( \sqrt{\kappa}+ \sqrt{\big(\frac{\omega}{n} + \sqrt{\frac{\omega}{n}}\big)\omega\kappa} + \omega\Big)\log \frac{1}{\epsilon}\right)$} \hspace{-2.5mm}
		& ---
		& \makecell{\yes compression \\ 
			\yes acceleration } \\ \hline
		
		\rowcolor{bgcolor}
		\gape{\makecell{\canita\\ (this paper)}}
		& ---
		& \hspace{-2mm} $O${\scriptsize $\left(\sqrt{\Big(1+\sqrt{\frac{\omega^3}{n}}\Big)\frac{L}{\epsilon}} + \omega\big(\frac{1}{\epsilon}\big)^{\frac{1}{3}}\right)$} \hspace{-2.5mm}
		&\gape{\makecell{\yes compression \\ 
				\yes acceleration}} \\ \hline
	\end{tabular}
\end{table}

\section{Summary of Contributions}
\label{sec:contrib}

In this paper we propose and analyze an accelerated gradient method with compressed communication, which we call \canita (described in Algorithm~\ref{alg:canita}), for solving distributed {\em general convex} optimization problems of the form \eqref{eq:prob}. In particular, \canita can loosely be seen as a combination of the accelerated gradient method \algname{ANITA} of \cite{li2021anita}, and the variance-reduced compressed gradient method \algname{DIANA} of \cite{DIANA}. Ours is the first work provably combining the benefits of communication compression and acceleration in the general convex regime. 



\subsection{First accelerated rate for compressed gradient methods in the convex regime} For general convex problems, \canita is the first compressed communication gradient method with an \emph{accelerated rate}. In particular, our \canita solves the distributed problem \eqref{eq:prob} in $$O\left(\sqrt{\bigg(1+\sqrt{\tfrac{\omega^3}{n}}\bigg)\tfrac{L}{\epsilon}} + \omega\left(\tfrac{1}{\epsilon}\right)^{\frac{1}{3}}\right)$$ communication rounds, which improves upon the current state-of-the-art result $$O\left( \left(1+\tfrac{\omega}{n}\right) \tfrac{L}{\epsilon} + \tfrac{\omega^2+n}{\omega+n}\tfrac{1}{\epsilon}\right)$$ achieved by the \algname{DIANA} method \citep{khaled2020unified}.  See Table~\ref{table:1} for more comparisons.

Let us now illustrate the improvements coming from this new bound on an example with concrete numerical values. Let the compression ratio be $10\%$ (the size of compressed message is  $0.1\cdot d$, where $d$ is the size of the uncompressed message). If random sparsification or quantization is used to achieve this, then $\omega \approx 10$  (see Section \ref{sec:compress}). Further, if the number of devices/machines is $n=10^6$, and the target error tolerance is $\epsilon=10^{-6}$, then the number of communication rounds of our \canita method is $O(10^3)$, while the number of communication rounds of the previous state-of-the-art method \algname{DIANA} \citep{khaled2020unified} is $O(10^6)$, i.e., $O\big(\sqrt{\frac{L}{\epsilon}}\big)$ vs.\  $O(\frac{L}{\epsilon})$. {\em This is an improvement of three orders of magnitude.}

Moreover, the numerical experiments in Section~\ref{sec:exp} indeed show that the performance of our \canita is much better than previous non-accelerated compressed methods (\algname{QSGD} and \algname{DIANA}), corroborating the theoretical results (see Table~\ref{table:1}) and confirming the practical superiority of our accelerated \canita method.

\subsection{Accelerated rate with limited compression for free} For strongly convex problems, \citet{li2020acceleration} showed that if the number of devices/machines $n$ is large, or the compression variance parameter $\omega$ is not very high ($\omega \leq n^{1/3}$), then their \algname{ADIANA} method enjoys the benefits of both compression and acceleration (i.e., $\sqrt{\kappa}\log\frac{1}{\epsilon}$ of \algname{ADIANA} vs.\  $\kappa\log\frac{1}{\epsilon}$ of previous works).   

In this paper, we consider the general convex setting and show that the proposed \canita also enjoys the benefits of both compression and acceleration. Similarly, if $\omega \leq n^{1/3}$ (i.e., many devices, or limited compression variance),  \canita achieves the accelerated rate $\sqrt{\frac{L}{\epsilon}}$ vs.\  $\frac{L}{\epsilon}$ of previous works.
This means that the compression does not hurt the accelerated rate at all.
Note that the second term $\big(\frac{1}{\epsilon}\big)^{\frac{1}{3}}$ is of a lower order compared with the first term $\sqrt{\frac{L}{\epsilon}}$.

\subsection{Novel proof technique} The proof behind the analysis of \canita is significantly different from that of \algname{ADIANA} \citep{li2020acceleration}, which critically relies on  strong convexity. Moreover, the theoretical rate in the strongly convex case is linear $O(\log\frac{1}{\epsilon})$, while it is sublinear $O(\frac{1}{\epsilon})$ or $O\big(\sqrt{\frac{1}{\epsilon}}\big)$ (accelerated) in the general convex case.  We hope that our novel analysis can provide new insights and shed light on future work.

\section{Preliminaries}

Let $[n]$ denote the set $\{1,2,\cdots,n\}$ and $\n{\cdot}$ denote the Euclidean norm for a vector and the spectral norm for a matrix.
Let $\inner{u}{v}$ denote the standard Euclidean inner product of two vectors $u$ and $v$.
We use $O(\cdot)$ and $\Omega(\cdot)$ to hide the absolute constants.

\subsection{Assumptions about the compression operators}
\label{sec:compress}

We now introduce the notion of a randomized {\em compression operator} which we use to compress the gradients to save on communication. We rely on a standard class of unbiased compressors (see Definition~\ref{def:comp}) that was used in the context of distributed gradient methods before \citep{alistarh2017qsgd, khirirat2018distributed, DIANA2, li2020unified, li2020acceleration}. 

\begin{definition}[Compression operator]\label{def:comp}
	A randomized map $\calC: \R^d\mapsto \R^d$ is an $\omega$-compression operator  if 
	\begin{equation}\label{eq:comp}
	\Exp{\calC(x)}=x, \qquad \Exp{\ns{\calC(x)-x}} \leq \omega\ns{x}, \qquad \forall x\in \R^d.
	\end{equation}
	In particular, no compression ($\calC(x)\equiv x$) implies $\omega=0$.
\end{definition}

It is well known that the conditions \eqref{eq:comp} are satisfied by many practically useful compression operators (see Table~1 in \citep{biased2020,UP2021}). For illustration purposes, we now present a couple canonical examples: sparsification and quantization.

\paragraph{Example 1 (Random sparsification).}
Given $x\in \R^d$, the random-$k$ sparsification operator is defined by $$\calC(x)\eqdef \frac{d}{k} \cdot (\xi_k \odot x),$$ where 
$\odot$ denotes the Hadamard (element-wise) product and $\xi_k\in \{0,1\}^d$ is a uniformly random binary vector with $k$ nonzero entries ($\n{\xi_k}_0=k$).
This random-$k$ sparsification operator $\calC$ satisfies \eqref{eq:comp} with $\omega =\frac{d}{k}-1$.
By setting $k=d$, this reduces to the identity compressor, whose variance is obviously zero: $\omega=0$.

\paragraph{Example 2 (Random quantization).}
Given $x\in \R^d$, the ($p,s$)-quantization operator  is defined by
$$\calC(x)\eqdef \text{sign}(x)\cdot \n{x}_p\cdot \frac{1}{s} \cdot \xi_s,$$ 
where $p,s\geq 1$ are integers, and $\xi_s\in\R^d$ is a random vector with 
$i$-th element $$\xi_s(i)\eqdef \begin{cases}
l+1, &\text {with probability } \frac{|x_i|}{\n{x}_p}s-l ,\\
l, &\text {otherwise.}
\end{cases}$$ The level $l$ satisfies $\frac{|x_i|}{\n{x}_p}\in [\frac{l}{s}, \frac{l+1}{s}]$. The probability is chosen so that $\Exp{\xi_s(i)} = \frac{|x_i|}{\n{x}_p}s$.
This ($p,s$)-quantization operator $\calC$ satisfies \eqref{eq:comp} with $\omega = 2+\frac{d^{1/p}+d^{1/2}}{s}$.
In particular, \algname{QSGD} \citep{alistarh2017qsgd} used $p=2$ (i.e., ($2,s$)-quantization) and proved that the expected sparsity of $\cC(x)$ is $\Exp{\n{\calC(x)}_0} = O\big(s(s+\sqrt{d})\big)$.

\subsection{Assumptions about the functions}

Throughout the paper, we assume that the functions $f_i$ are convex and have Lipschitz continuous gradient.

\begin{assumption}\label{asp:smooth}
	Functions $f_i:\R^d\to \R$ are convex, differentiable, and $L$-smooth. The last condition means that there exists a constant $L>0$ such that for  all $i\in[n]$ we have
	\begin{equation}\label{eq:smooth}
	\n{\nabla f_i(x) - \nabla f_i(y)}\leq L \n{x-y}, \qquad \forall x,y\in \R^d.
	\end{equation}
	
\end{assumption}

It is easy to see that the objective $f(x)= \frac{1}{n}\sum_{i=1}^n{f_i(x)}$ in \eqref{eq:prob}  satisfies \eqref{eq:smooth} provided that the constituent functions $\{f_i\}$ do.

\section{The \canita Algorithm}
\label{sec:alg}

\begin{algorithm}[b]
	\caption{Distributed compressed accelerated \algname{ANITA} method (\canita)}
	\label{alg:canita}
	\begin{algorithmic}[1]
		\REQUIRE 
		initial point $x^0\in \R^d$, initial shift vectors $h_1^0,\dots, h_n^0 \in \R^d$, probabilities $\{\pt\}$,  and positive stepsizes $\{\alphat\}, \{\etat\}, \{\thetat\}$ \\
		\STATE {\bf Initialize:} $w^0=\bx^0=x^0$ and $h^0=\frac{1}{n}\sum_{i=1}^{n}h_i^0$   
		\FOR{$t = 0,1,2,\ldots$}
		\STATE $\uxt = \thetat \xt + (1 - \thetat)\wt$ \label{line:uxt}
		\STATE {\bf{for all machines $i= 1,2,\ldots,n$ do in parallel}}
		\STATE \quad Compress the shifted local gradient $\cit(\nabla f_i(\uxt) - \hit)$ and send the result to the server
		\label{line:shiftgrad}
		\STATE \quad Update the local shift $\hitn= \hit +\alpha_t \cit(\nabla f_i(\wt) - \hit)$ \label{line:shifth}
		\STATE {\bf{end for}}
		\STATE Aggregate received compressed local gradient information:\\
		\qquad $\gt = \allht + \frac{1}{n}\sum \limits_{i=1}^n \cit(\nabla f_i(\uxt) - \hit) $ \hfill $\bullet$ Compute gradient estimator \\ \label{line:aggregategrad}
		\qquad $\htn = \allht + \alpha_t  \frac{1}{n}\sum \limits_{i=1}^n \cit(\nabla f_i(\wt) - \hit)$  \hfill $\bullet$ Maintain the average of local shifts  
		\STATE Perform update step:\\ 
		\qquad $\xtn = \xt - \frac{\etat}{\thetat} \gt$  \label{line:update}
		\STATE $\bxtn = \thetat \xtn + (1-\thetat) \wt$
		\label{line:bxtn}
		\STATE $\wtn = \begin{cases}
		\bxtn, &\text {with probability } \pt\\
		\wt, &\text {with probability } 1-\pt
		\end{cases}$ \label{line:prob} 
		\ENDFOR
	\end{algorithmic}
\end{algorithm}

In this section, we describe our method, for which we coin the name \canita, designed for solving problem \eqref{eq:prob}, which is of importance  in distributed and federated learning, and contrast it to the most closely related methods \algname{ANITA}~\citep{li2021anita},   \algname{DIANA}~\citep{DIANA} and \algname{ADIANA}~\citep{li2020acceleration}.

\subsection{\canita: description of the method}

Our proposed method \canita, formally described in Algorithm~\ref{alg:canita}, is an accelerated gradient method supporting compressed communication. It is the first  method combing the benefits of acceleration and compression in the general convex regime (without strong convexity).

In each round $t$,  each machine computes its local gradient (e.g., $\nabla f_i(\uxt)$)  and then a shifted version is
compressed and sent to the server (See Line~\ref{line:shiftgrad} of Algorithm~\ref{alg:canita}).
The local shifts $\hit$ are adaptively changing throughout the iterative process (Line~\ref{line:shifth}), and have the role of reducing the variance introduced by compression $\cC(\cdot)$.  If no compression is used, we may simply set the shifts to be $\hit=0$ for all $i, t$.
The server subsequently aggregates all received messages to obtain the gradient estimator $\gt$ and maintain the average of local shifts $\htn$ (Line~\ref{line:aggregategrad}), and then 
perform gradient update step (Line~\ref{line:update}) and update momentum sequences (Line~\ref{line:bxtn} and \ref{line:uxt}).
Besides, the last Line~\ref{line:prob} adopts a randomized update rule for the auxiliary vectors $\wt$ which simplifies the algorithm and analysis, resembling the workings of the loopless SVRG method used in \citep{L-SVRG,li2021anita}.

\subsection{\canita vs existing methods}

\canita can be loosely seen as a combination of  the accelerated gradient method \algname{ANITA} of \cite{li2021anita}, and  the variance-reduced compressed gradient method \algname{DIANA} of \cite{DIANA}.
In particular, \canita uses momentum/acceleration steps (see Line~\ref{line:uxt} and \ref{line:bxtn} of Algorithm~\ref{alg:canita}) inspired by those of \algname{ANITA} \citep{li2021anita}, and adopts the shifted compression framework for each machine (see Line~\ref{line:shiftgrad} and \ref{line:shifth} of Algorithm~\ref{alg:canita}) as in the \algname{DIANA} method \citep{DIANA}.  \begin{quote} \em We prove that \canita enjoys the benefits of both methods  simultaneously, i.e., convergence acceleration of \algname{ANITA} and gradient compression of \algname{DIANA}.
\end{quote}

Although \canita can conceptually be seen as  combination of \algname{ANITA} \citep{li2021anita} and \algname{DIANA} \citep{DIANA, DIANA2, khaled2020unified} from an algorithmic perspective,  the analysis of \canita is entirely different.  Let us now briefly outline some of the main differences.
\begin{itemize}
	\item For example, compared with \algname{ANITA} \citep{li2021anita}, \canita needs to deal with the extra compression of shifted local gradients in the distributed network. Thus, the obtained gradient estimator $g^k$ in Line~\ref{line:aggregategrad} of Algorithm~\ref{alg:canita} is substantially different and more complicated than the one in \algname{ANITA}, which necessitates a novel proof technique.
	\item Compared with \algname{DIANA} \citep{DIANA, DIANA2, khaled2020unified}, the extra momentum steps in Line~\ref{line:uxt} and \ref{line:bxtn} of Algorithm~\ref{alg:canita} make the analysis of \canita more complicated than that of \algname{DIANA}. We obtain the accelerated rate $O\big(\sqrt{\frac{L}{\epsilon}}\big)$ rather than  the non-accelerated rate $O(\frac{L}{\epsilon})$ of \algname{DIANA}, and this is impossible without a substantially different proof technique. 
	\item Compared with the accelerated \algname{DIANA} method \algname{ADIANA} of \cite{li2020acceleration},
	the analysis of \canita is also substantially different since \canita cannot exploit the strong convexity assumed therein. \end{itemize}

Finally, please refer to Section~\ref{sec:contrib} where we summarize our contributions for additional discussions.

\section{Convergence Results for the \canita Algorithm}

In this section, we provide  convergence results for  \canita (Algorithm~\ref{alg:canita}).
In order to simplify the expressions appearing in our  main result (see Theorem~\ref{thm:main} in Section~\ref{subsec:main}) and in the lemmas needed to prove it (see Appendix~\ref{sec:proofsketch}), it will be convenient to let
\begin{equation}\label{eq:F,H,D}
\vspace{-2mm}
F^t \eqdef f(\wt)-f(x^*), 
\qquad H^t \eqdef  \frac{1}{n}\cht, 
\qquad D^t \eqdef  \frac{1}{2}\ns{x^t-x^*}.
\end{equation}

\subsection{Generic convergence result} \label{subsec:main}
We first present the main convergence theorem of \canita for solving the distributed optimization problem \eqref{eq:prob} in the general convex regime. 

\begin{theorem}\label{thm:main}
	Suppose that Assumption~\ref{asp:smooth} holds and the compression operators $\{\cC_i^t\}$ used in Algorithm~\ref{alg:canita} satisfy \eqref{eq:comp} of Definition~\ref{def:comp}.
	For any two positive sequences $\{\beta_t\}$ and $\{\gamma_t\}$ such that the probabilities $\{p_t\}$ and positive stepsizes $\{\alphat\}, \{\etat\}, \{\thetat\}$ of Algorithm~\ref{alg:canita} satisfy the following relations \begin{equation}\label{eq:two}
	\small
	\alphat\leq \frac{1}{1+\omega}, 
	\qquad \etat \leq \frac{1}{L\left(1+\beta_t+4\pt\gammat \big(1+ \frac{2\pt}{\alphat} \big) \right)} 
	\end{equation} 
	for all $t\geq 0$, and  
	\begin{equation}\label{eq:parat1}
	\small
	\frac{2\omega}{\beta_t n}
	+ 4\pt\gamma_t \Big(1 + \frac{2\pt}{\alphat}\Big) \leq 1-\thetat,
	\quad \frac{(1-p_t\theta_t)\eta_t}{p_t\theta_t^2} \leq \frac{\eta_{t-1}}{p_{t-1}\theta_{t-1}^2}, 
	\quad \left(\frac{\omega}{\betat n}+\Big(1-\frac{\alphat}{2}\Big)\gammat\right)\frac{\etat}{\thetat^2} \leq \frac{\gamma_{t-1}\eta_{t-1}}{\theta_{t-1}^2}
	\end{equation} 
	for all $t\geq1$. Then the sequences $\{\xt, \wt,  \hit\}$ of \canita (Algorithm~\ref{alg:canita}) for all $t\geq 0$ satisfy the inequality
	\begin{align}
	\EB{F^{t+1} + \frac{\gammat\pt}{ L} H^{t+1}} \leq \frac{\thetat^2\pt}{\etat}\left(\frac{(1-\theta_0p_0)\eta_0}{\theta_0^2p_0}F^0	+ \Big(\frac{\omega}{\beta_0 n}+\Big(1-\frac{\alpha_0}{2}\Big)\gamma_0\Big)\frac{\eta_0}{\theta_0^2 L}H^0	 
	+ D^0
	\right),  \label{eq:thm-main}
	\end{align}
	where the quantities $F^t,H^t,D^t$ are defined in \eqref{eq:F,H,D}.	
\end{theorem}

The detailed proof of Theorem~\ref{thm:main} which relies on six lemmas is provided in Appendix~\ref{sec:proofsketch}. 
In particular, the proof simply follows from the key Lemma~\ref{lem:key-main} (see Appendix~\ref{sec:proofofthm1}), while Lemma~\ref{lem:key-main} closely relies on previous five Lemmas~\ref{lem:func-main}--\ref{lem:fwfx-main} (see Appendix~\ref{app:proofofkeylem}). 
Note that all proofs for these six lemmas are deferred to Appendix~\ref{sec:proofsketch-appendix}.

As we shall see in detail in Section~\ref{subsec:detailrate}, the sequences $\beta_t,\gamma_t, p_t$ and $\alpha_t$ can be fixed to some constants.\footnote{Exception: While we indeed choose $\beta_t\equiv \beta$ for $t\geq 1$, the value of $\beta_0$ may be different.}
However, the relaxation parameter $\theta_t$ needs to be decreasing and the stepsize $\eta_t$ may be increasing until a certain threshold. In particular, we choose  
\begin{align}
\beta_t \equiv c_1, \quad \gamma_t \equiv c_2,  \quad  p_t\equiv c_3, \quad \alpha_t \equiv c_4,  \quad \theta_t=\frac{c_5}{t+c_6}, \quad  \eta_t =\min\Big\{\Big(1+\frac{1}{t+c_7}\Big)\eta_{t-1},~ \frac{1}{c_8L}\Big\}, \label{eq:simplepara}
\end{align} 
where the constants $\{c_i\}$ may depend on the compression parameter $\omega$ and the number of devices/machines $n$. 
As a result, the right hand side of \eqref{eq:thm-main} will be of the order $O\left(\frac{L}{t^2}\right)$, which indicates an {\em accelerated} rate.  Hence, in order to find an $\epsilon$-solution of problem \eqref{eq:prob}, i.e., vector $w^{T+1}$ such that
\begin{equation}\label{eq:g987g9gd9gvcbu9v8}
\Exp{f(w^{T+1}) - f(x^*)} \overset{\eqref{eq:F,H,D}}{\eqdef}\Exp{F^{T+1}}  \leq \epsilon,
\end{equation}
the number of communication rounds of \canita (Algorithm~\ref{alg:canita}) is at most
$
T= O\Big(\sqrt{\frac{L}{\epsilon}}\Big).
$

While the above rate has an accelerated dependence on $\epsilon$, it will be crucial to study the omitted constants $\{c_i\}$ (see \eqref{eq:simplepara}), and in particular their dependence on the compression parameter $\omega$ and the number of devices/machines $n$. As expected, for any fixed target error $\epsilon>0$, the number of communication rounds $T$ (sufficient to guarantee that \eqref{eq:g987g9gd9gvcbu9v8} holds) may grow with increasing levels of compression, i.e., with increasing $\omega$. However, at the same time, the communication cost in each round decreases with $\omega$.  It is easy to see that this trade-off benefits compression. 
In particular, as we mention in Section~\ref{sec:contrib}, 
if the number of devices $n$ is large, or the compression variance $\omega$ is not very high, then compression does not hurt the accelerated rate of communication rounds at all.

\subsection{Detailed convergence result} 
\label{subsec:detailrate}

We now formulate a concrete Theorem~\ref{thm:detail} from Theorem~\ref{thm:main} which leads to a detailed convergence result for \canita (Algorithm~\ref{alg:canita}) by specifying the choice of the parameters $\beta_t,\gamma_t, p_t, \alpha_t, \theta_t$ and $\eta_t$.
The detailed proof of Theorem~\ref{thm:detail} is deferred to Appendix~\ref{app:proofofthm2}.

\begin{theorem}\label{thm:detail}
	Suppose that Assumption \ref{asp:smooth} holds and the compression operators $\{\cC_i^t\}$ used in Algorithm~\ref{alg:canita} satisfy \eqref{eq:comp} of Definition~\ref{def:comp}.
	Let $b=\min\Big\{\omega, \sqrt{\frac{\omega(1+\omega)^2}{n}}\Big\}$ and choose the two positive sequences $\{\beta_t\}$ and $\{\gamma_t\}$ as follows:
	\begin{align} \label{eq:set-beta}	
	\betat =\begin{cases} 
	\beta_0=\frac{9(1+b+\omega)^2}{(1+b)L} & \text{for}~~ t=0 \\
	\beta \equiv  \frac{48\omega(1+\omega)(1+b+2(1+\omega))}{n(1+b)^2} & \text{for}~~  t\geq 1
	\end{cases}, \qquad 
	\gammat =\gamma \equiv  \frac{(1+b)^2}{8(1+b+2(1+\omega))} \quad \text{for}~~  t\geq 0.
	\end{align}
	If we set the probabilities $\{p_t\}$ and positive stepsizes $\{\alphat\}, \{\etat\}, \{\thetat\}$ of Algorithm~\ref{alg:canita} as follows:
	\begin{align} \label{eq:set-parameter}	
	\pt \equiv \frac{1}{1+b}, 
	\qquad \alphat \equiv \frac{1}{1+\omega}, 
	\qquad \thetat = \frac{3(1+b)}{t+9(1+b+\omega)}, \quad \text{for}~~ t\geq 0,
	\end{align}
	and 
	\begin{align} \label{eq:set-eta}	
	\etat =\begin{cases} 
	\frac{1}{L(\beta_0 + 3/2)} & \text{for}~~ t=0 \\
	\min\Big\{\big(1+\frac{1}{t+9(1+b+\omega)}\big)\eta_{t-1},~ \frac{1}{L(\beta+3/2)}\Big\} & \text{for}~~  t\geq 1
	\end{cases}.
	\end{align}
	Then  \canita (Algorithm~\ref{alg:canita}) for all  $T\geq 0$ satisfies 	
	\begin{align}
	\EB{F^{T+1}} \leq   O\left(\frac{(1+\sqrt{\omega^3/n})L}{T^2} + \frac{\omega^3}{T^3}\right). \label{eq:cor}
	\end{align}
	According to \eqref{eq:cor}, the number of communication rounds for \canita (Algorithm~\ref{alg:canita}) to find an $\epsilon$-solution of the distributed problem \eqref{eq:prob}, i.e., 	
	$$ 
	\Exp{f(w^{T+1}) - f(x^*)} \overset{\eqref{eq:F,H,D}}{\eqdef}\Exp{F^{T+1}}  \leq \epsilon,
	$$
	is at most 
	$$
	T= O\left(\sqrt{\bigg(1+\sqrt{\frac{\omega^3}{n}}\bigg)\frac{L}{\epsilon}}
	+ \omega\left(\frac{1}{\epsilon}\right)^{\frac{1}{3}} \right).
	$$
\end{theorem}

\section{Experiments}
\label{sec:exp}

In this section, we demonstrate the performance of our accelerated method \canita (Algorithm~\ref{alg:canita}) and previous methods \algname{QSGD} and \algname{DIANA} (the theoretical convergence results of these algorithms can be found in Table~\ref{table:1}) with different compression operators on the logistic regression problem, 
\begin{equation}\label{eq:prob-exp}
\min_{x\in \R^d}   f(x):= \frac{1}{n}\sum_{i=1}^n \log\big(1+\exp(-b_ia_i^Tx)\big),
\end{equation}
where $\{a_i,b_i\}_{i=1}^n\in \R^{d}\times \{\pm 1\}$ are data samples.
We use three standard datasets: \texttt{a9a}, \texttt{mushrooms}, and \texttt{w8a} in the experiments.
All datasets are downloaded from LIBSVM \citep{chang2011libsvm}.

Similar to \citet{li2020acceleration}, we also use three different compression operators: \emph{random sparsification} (e.g.~\citep{stich2018sparsified}), \emph{natural compression} (e.g.~\citep{Cnat}), and \emph{random quantization} (e.g.~\citep{alistarh2017qsgd}). 
In particular, we follow the same settings as in \citet{li2020acceleration}.
For random-$r$ sparsification, the number of communicated bits per iteration is $32r$,  and we choose $r = d/4$. For natural compression, the number of communicated bits  per iteration is $9d$ bits \citep{Cnat}. 
For random $(2,s)$-quantization, we choose $s = \sqrt{d}$, which means the number of communicated bits per iteration is $2.8d+32$ \citep{alistarh2017qsgd}. 
The default number of nodes/machines/workers is $20$.
In our experiments, we directly use the theoretical stepsizes and parameters for all three algorithms: \algname{QSGD}~\citep{alistarh2017qsgd, li2020unified}, \algname{DIANA}~\citep{khaled2020unified}, our \canita (Algorithm~\ref{alg:canita}). 
To compare with the settings of \algname{DIANA} and \canita, we use local gradients (not stochastic gradients) in \algname{QSGD}. Thus here \algname{QSGD} is equivalent to \algname{DC-GD} provided in \cite{li2020unified}.

\begin{figure}[t]
	\centering
		\includegraphics[width=0.325\linewidth]{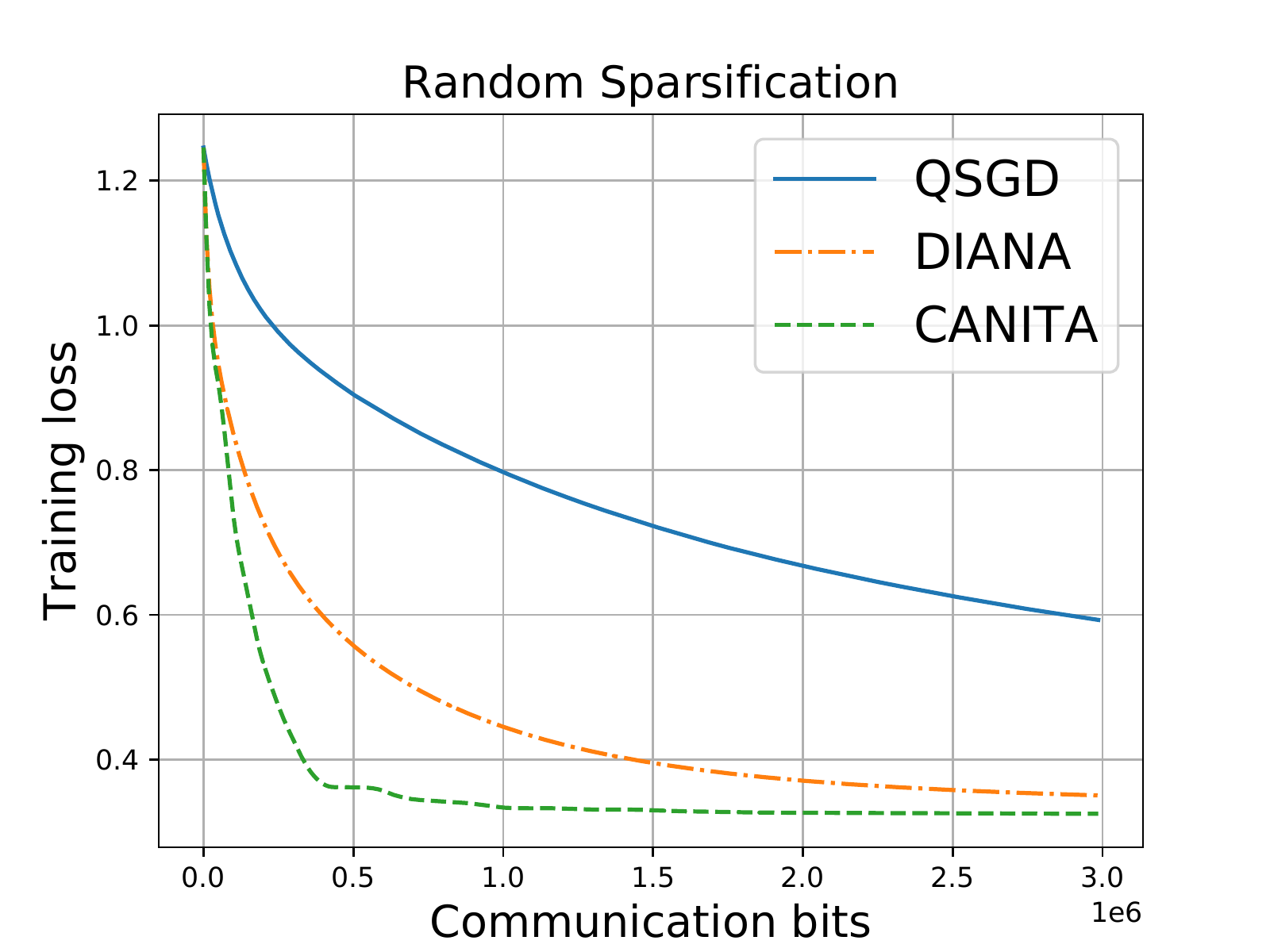}
		\includegraphics[width=0.325\linewidth]{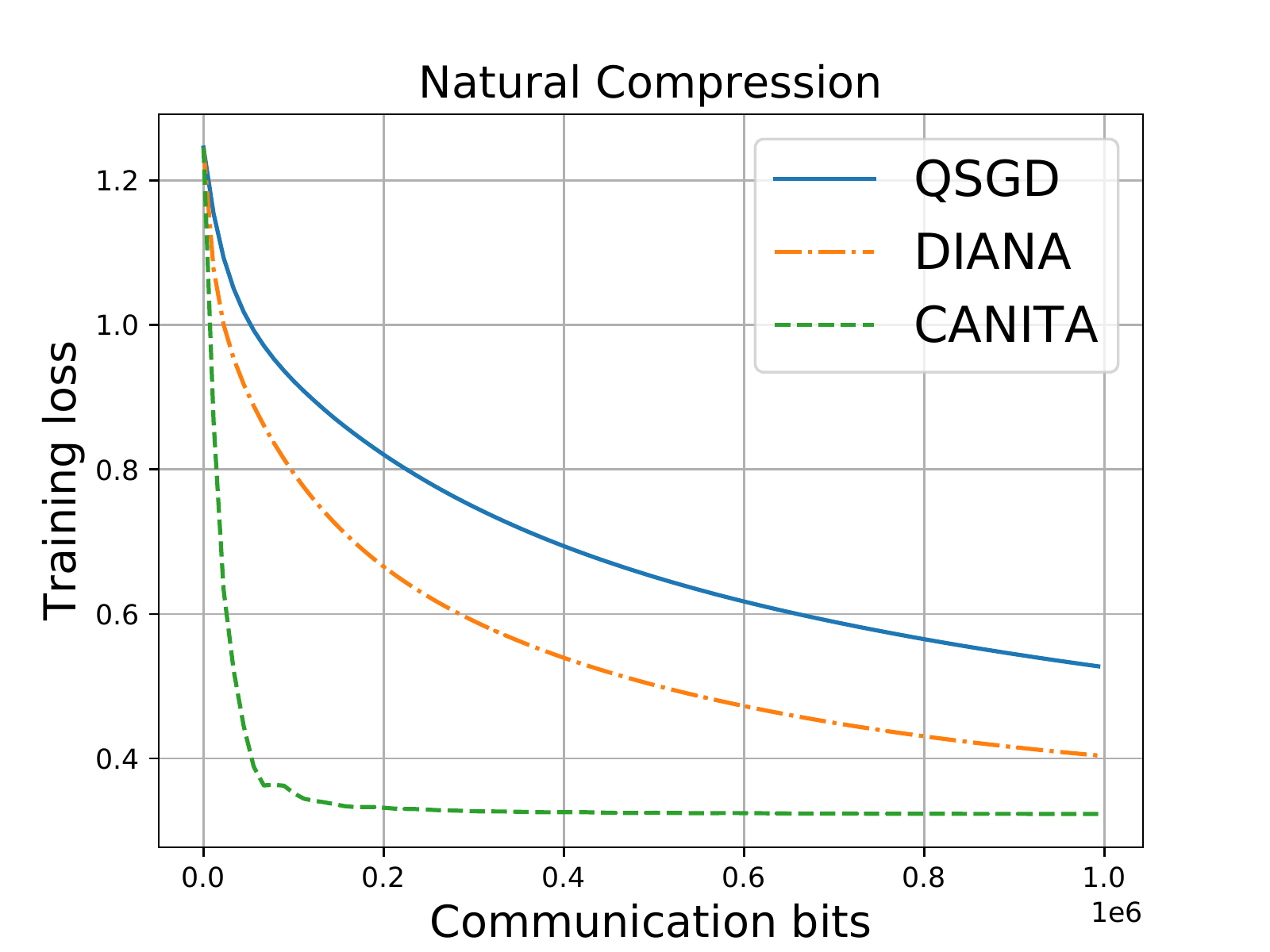}
		\includegraphics[width=0.325\linewidth]{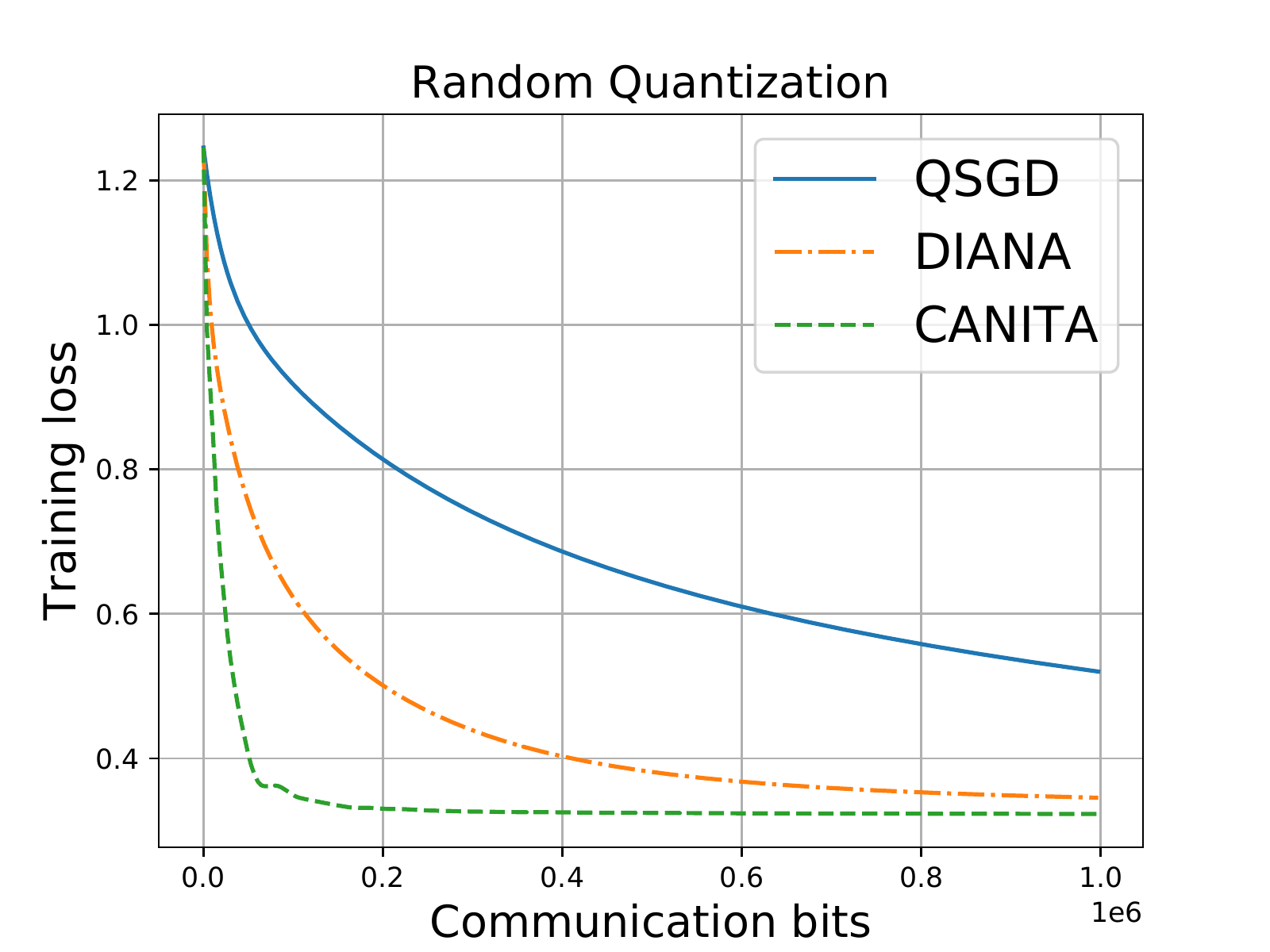}
	\caption{Performance of different methods for three different compressors (random sparsification, natural compression, and random quantization) on the \texttt{a9a} dataset.}
	\label{fig:a9a}
	\vspace{4mm}
	\centering
	\includegraphics[width=0.325\linewidth]{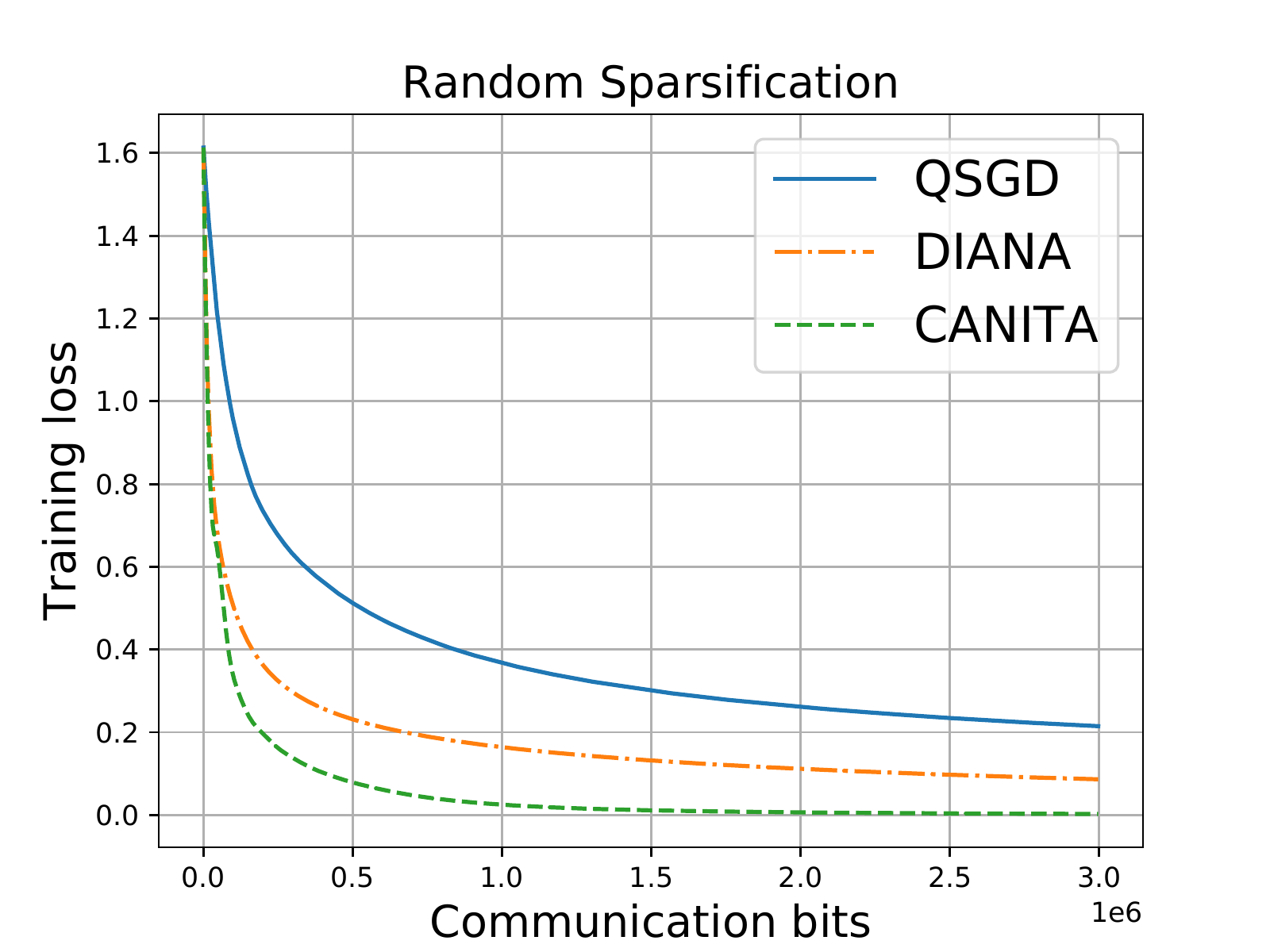}
	\includegraphics[width=0.325\linewidth]{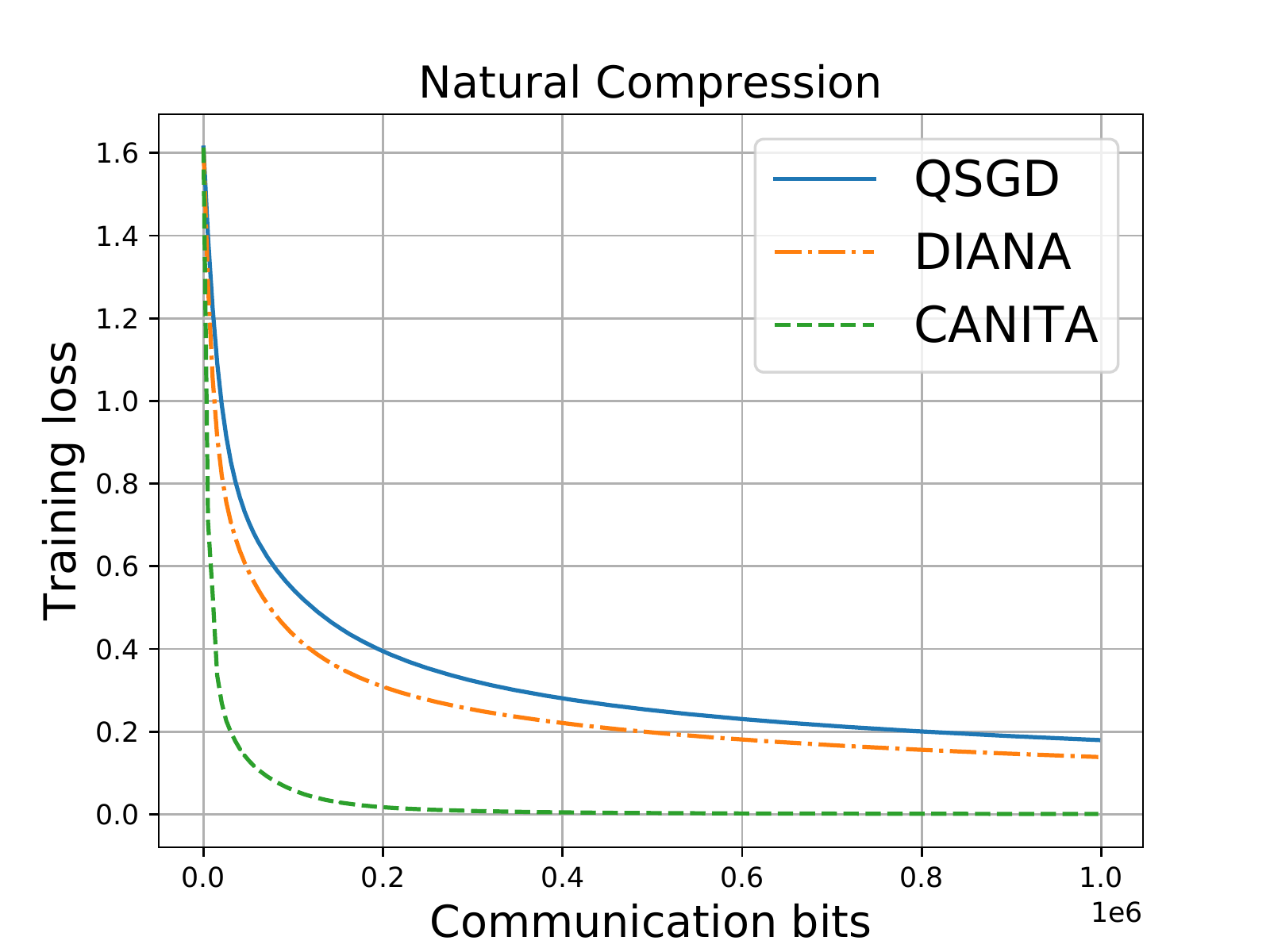}
	\includegraphics[width=0.325\linewidth]{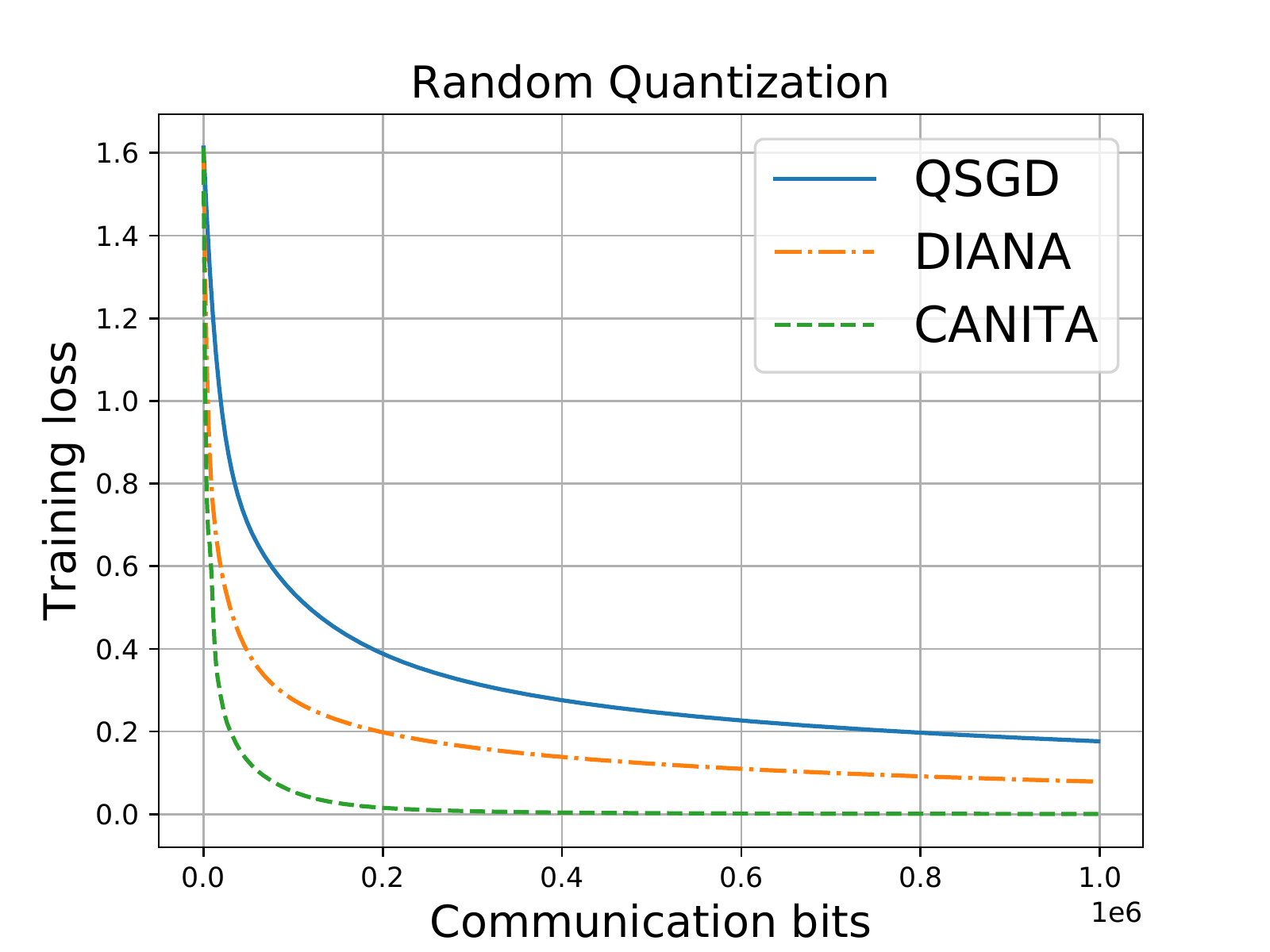}
	\caption{Performance of different methods for three different compressors (random sparsification, natural compression, and random quantization) on the \texttt{mushrooms} dataset.}
	\label{fig:mushrooms}
	\vspace{4mm}
	\centering
	\includegraphics[width=0.325\linewidth]{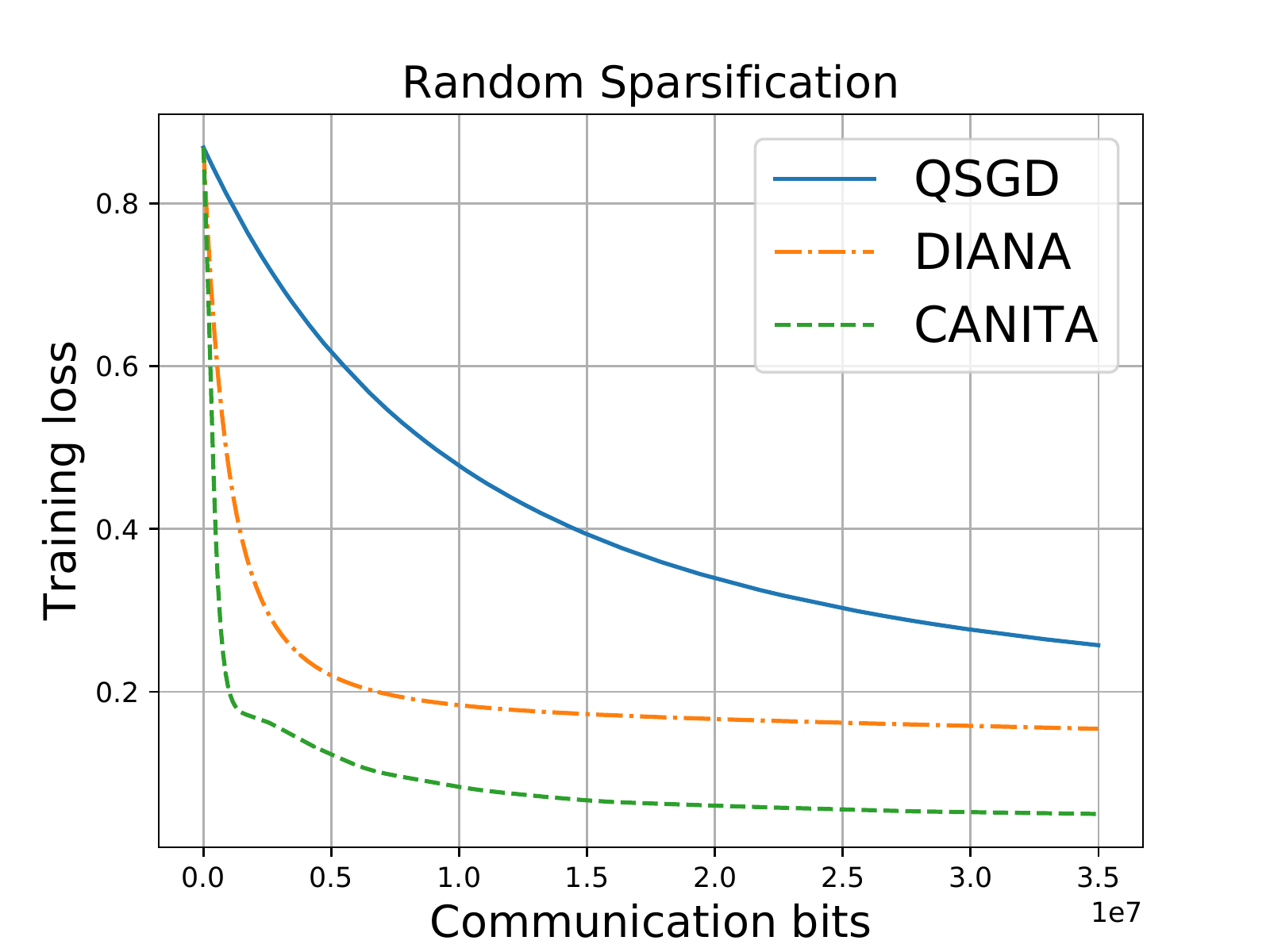}
	\includegraphics[width=0.325\linewidth]{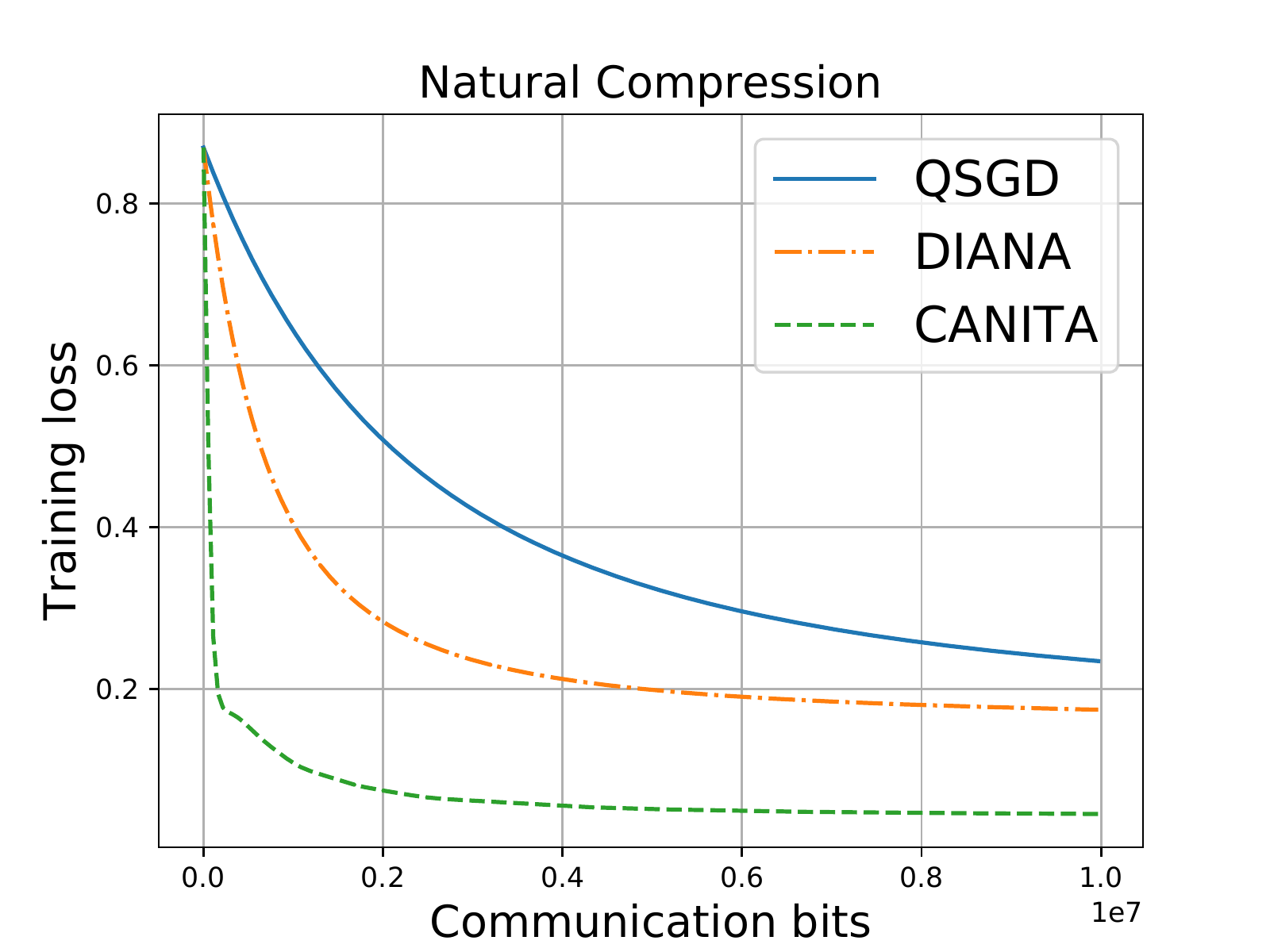}
	\includegraphics[width=0.325\linewidth]{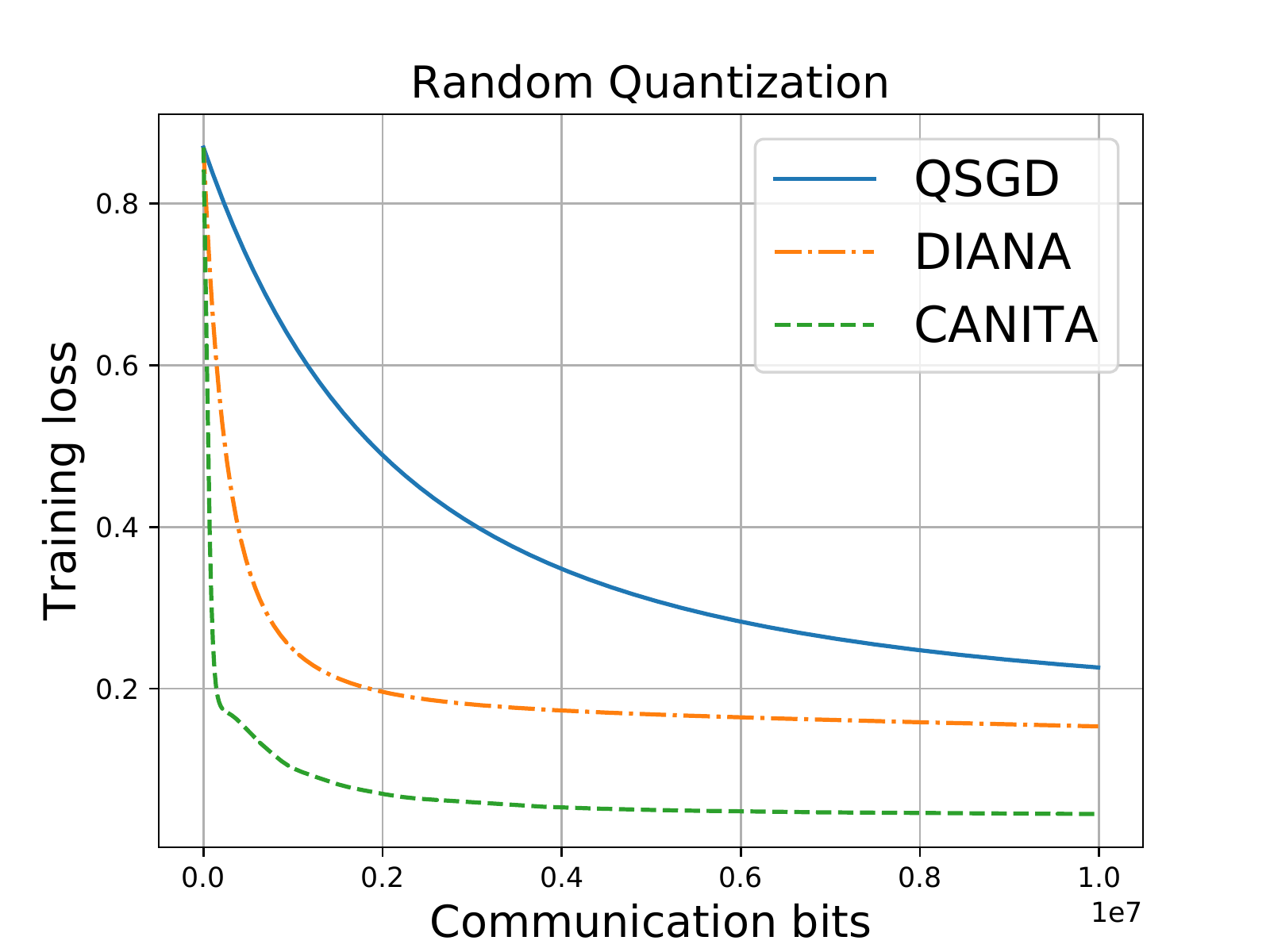}
	\caption{Performance of different methods for three different compressors (random sparsification, natural compression, and random quantization) on the \texttt{w8a} dataset.}
	\label{fig:w8a}
\end{figure}

In Figures~\ref{fig:a9a}--\ref{fig:w8a}, we compare our \canita with \algname{QSGD} and \algname{DIANA} with three compression operators: random sparsification (left), natural compression (middle), and random quantization (right) on three datasets: \texttt{a9a} (Figure~\ref{fig:a9a}), \texttt{mushrooms} (Figure~\ref{fig:mushrooms}), and \texttt{w8a} (Figure~\ref{fig:w8a}).
The $x$-axis and $y$-axis represent the number of communication bits and the training loss, respectively.

Regarding the different compression operators, the experimental results indicate that natural compression and random quantization are better than random sparsification for all three algorithms. For instance, in Figure~\ref{fig:a9a}, \algname{DIANA} uses $1.5\times 10^6$ (random sparsification),  $1.0\times 10^6$ (natural compression),  $0.4\times 10^6$ (random quantization) communication bits for achieving the loss $0.4$, respectively.

Moreover, regarding the different algorithms, the experimental results indeed show that our \canita converges the fastest compared with both \algname{QSGD} and \algname{DIANA} for all three compressors in all Figures~\ref{fig:a9a}--\ref{fig:w8a},
validating the theoretical results (see Table~\ref{table:1}) and confirming the practical superiority of our accelerated \canita method.

\section{Conclusion}
In this paper, we proposed \canita: the first gradient method for distributed {\em general convex} optimization provably enjoying the benefits of both {\em communication compression} and {\em convergence acceleration}. There is very limited work on combing compression and acceleration. Indeed, previous works only focus on the (much simpler) strongly convex setting. We hope that our novel algorithm and analysis can provide new insights and shed light on future work in this line of research. We leave further improvements to future work. For example, one may ask whether our approach can be combined with the benefits provided by multiple local update steps \citep{FL2017-AISTATS, localSGD-Stich, localSGD-AISTATS2020, SCAFFOLD, FedPAGE}, with additional variance reduction techniques \citep{DIANA2, li2020unified}, and to what extent one can extend our results to structured nonconvex problems \citep{li2018simple, li2019ssrgd, li2021page, li2021short, li2021zerosarah, gorbunov2021marina, richtarik2021ef21, fatkhullin2021ef21}.

%

\bibliographystyle{plainnat}
\bibliography{canita}

\newpage
\appendix

\tableofcontents

\newpage

\section{Missing Proof for Theorem~\ref{thm:main} in Section~\ref{subsec:main}}
\label{sec:proofsketch}
In order to prove Theorem~\ref{thm:main}, we first formulate six auxiliary results (Lemmas~\ref{lem:func-main}--\ref{lem:key-main}) in Appendix~\ref{sec:lemmas}. The detailed proofs of these lemmas are deferred to Appendix~\ref{sec:proofsketch-appendix}. Then in Appendix~\ref{sec:proofofthm1} we show that Theorem~\ref{thm:main} follows from Lemma~\ref{lem:key-main}. 

\subsection{Six lemmas}
\label{sec:lemmas}

First, we need a useful Lemma~\ref{lem:func-main} which captures  the change of the function value after a single gradient update step.

\begin{lemma}\label{lem:func-main}
	Suppose that Assumption \ref{asp:smooth} holds. For any $\beta_t>0$, the following equation holds for \canita (Algorithm~\ref{alg:canita}) for any round $t\geq 0$:	
	\begin{align}
	\Exp{f(\bxtn)} &\leq \E\bigg[f(\uxt) +\inner{\nabla f(\uxt)}{\theta_t(x^*-\xt)} +\frac{\theta_t^2}{\etat}\left(D^t-D^{t+1} \right)	 \notag\\
	& \qquad \qquad
	-\left(\frac{\theta_t^2}{2\etat}-\frac{L(1+\beta_t)\theta_t^2}{2}\right)\ns{\xtn-\xt} 
	+\frac{1}{2L\beta_t}\ns{\nabla f(\uxt) - \tnabla} 
	\bigg]. \label{eq:func-main}
	\end{align}
\end{lemma}
Note that 
$$\bxtn - \uxt = \thetat (\xtn - \xt) = -\etat \gt$$ according to the two momentum/interpolation steps of \canita (see Line~\ref{line:uxt} and Line~\ref{line:bxtn} of Algorithm~\ref{alg:canita}) and the gradient update step (see Line~\ref{line:update} of Algorithm~\ref{alg:canita}).
The proof of Lemma~\ref{lem:func-main} uses these relations and the smoothness Assumption~\ref{asp:smooth}.

In the next lemma, we bound the last variance term $\Exp{\ns{\nabla f(\uxt) - \tnabla}}$ appearing in  \eqref{eq:func-main} of Lemma~\ref{lem:func-main}. To simplify the notation, from now on we will write
\begin{align}\label{eq:defYt}
Y^t \eqdef \frac{1}{n}\sum_{i=1}^n \ns{\nabla f_i(\wt)-\nabla f_i(\uxt)},
\end{align}
and recall that $H^t \eqdef  \frac{1}{n}\cht$ defined in \eqref{eq:F,H,D}. 
\begin{lemma}\label{lem:var-main}
	If $g^t$ is as defined in Line~\ref{line:aggregategrad} of Algorithm~\ref{alg:canita}, and the compression operator $\cC_i^t$ satisfies \eqref{eq:comp} of Definition~\ref{def:comp}, we have
	\begin{align}
	\Exp{\ns{\nabla f(\uxt) -\tnabla}}  \leq
	\frac{2\omega}{n} \left(Y^t	
	+ H^t \right).   \label{eq:var-main}
	\end{align}
\end{lemma}
This lemma is proved by using the definition of the $\omega$-compression operator (i.e., \eqref{eq:comp}).

Now, we need to bound the terms $Y^t$ and $H^t$ in \eqref{eq:var-main} of Lemma~\ref{lem:var-main}. 
We first show how to handle the term $H^t$ in the following Lemma~\ref{lem:shift-main}.

\begin{lemma}\label{lem:shift-main}
	Suppose that Assumption \ref{asp:smooth} holds and let $\alphat \leq \frac{1}{1+\omega}$. According to the probabilistic update of $\wtn$ in Line~\ref{line:prob} of Algorithm~\ref{alg:canita}, we have
	\begin{align}
	\EB{H^{t+1}}  &\leq \left(1-\frac{\alphat}{2}\right)H^t 
	+2\pt \left(1 + \frac{2\pt}{\alphat} \right) Y^t 
	+2\pt L^2\thetat^2\left(1 + \frac{2\pt}{\alphat} \right)\Exp{\ns{\xtn -\xt}}.  \label{eq:shift-main}
	\end{align}
\end{lemma}
This lemma is proved by using the update of $\wtn$ (Line~\ref{line:prob} of Algorithm~\ref{alg:canita}) and $\hitn$ (Line~\ref{line:shifth} of Algorithm~\ref{alg:canita}), the  property of $\omega$-compression operator (i.e., \eqref{eq:comp}), and the smoothness Assumption~\ref{asp:smooth}.

To deal with the term $Y^t$ in Lemmas~\ref{lem:var-main} and \ref{lem:shift-main}, we need the following result.

\begin{lemma}\label{lem:gradfunc-main}
	Suppose that Assumption \ref{asp:smooth} holds. For any $\uxt, \wt\in \R^d$, the following inequality holds:
	\begin{align}
	Y^t \leq  2L\Big(f(\wt) - f(\uxt) - \inner{\nabla f(\uxt)}{\wt-\uxt}\Big). \label{eq:gradfunc-main}
	\end{align}
\end{lemma}

The proof of this lemma directly follows from a standard result characterizing the $L$-smoothness of convex functions.

Finally, we also need a result connecting the function values $f(\bxtn)$ in \eqref{eq:func-main} of Lemma~\ref{lem:func-main} and $f(\wtn)$ in \eqref{eq:thm-main} of Theorem~\ref{thm:main} (recall that $F^{t+1} \eqdef f(\wtn)-f(x^*)$ in \eqref{eq:F,H,D}). 

\begin{lemma}\label{lem:fwfx-main}
	According to the probabilistic update of $\wtn$ in Line~\ref{line:prob} of Algorithm~\ref{alg:canita}, we have
	\begin{align}\label{eq:fwfx-main}
	\E[f(\wtn)] = \pt\E[f(\bxtn)] + (1-\pt)\E[f(\wt)].
	\end{align}
\end{lemma}

Now, we combine Lemmas~\ref{lem:func-main}--\ref{lem:fwfx-main} to obtain our final key lemma, which describes the recursive form of the objective function value after a single round.
\begin{lemma}\label{lem:key-main}
	Suppose that Assumption~\ref{asp:smooth} holds and the compression operators $\{\cC_i^t\}$ used in Algorithm~\ref{alg:canita} satisfy \eqref{eq:comp} of Definition~\ref{def:comp}.
	For any two positive sequences $\{\beta_t\}$ and $\{\gamma_t\}$ such that the probabilities $\{p_t\}$ and positive stepsizes $\{\alphat\}, \{\etat\}, \{\thetat\}$ of Algorithm~\ref{alg:canita} satisfy the following relations \begin{equation}\label{eq:twoXXX}
	\small
	\alphat\leq \frac{1}{1+\omega}, 
	\qquad \etat \leq \frac{1}{L\left(1+\beta_t+4\pt\gammat \big(1+ \frac{2\pt}{\alphat} \big) \right)} 
	\end{equation} for all $t\geq 0$, and
	\begin{equation}\label{eq:twoYYY} 
	\small
	\frac{2\omega}{\beta_t n}
	+ 4\pt\gamma_t \Big(1 + \frac{2\pt}{\alphat}\Big) \leq 1-\thetat
	\end{equation}	
	for all $t\geq 1$.
	Then the sequences $\{\xt, \wt,  \hit\}$ of \canita (Algorithm~\ref{alg:canita}) for all $t\geq 0$ satisfy the inequality	
	\begin{align}
	\EB{F^{t+1} + \frac{\gammat\pt}{L}H^{t+1}} \leq \E\bigg[(1-\thetat\pt)F^t
	+  \Big(\frac{\omega}{\betat n}+\Big(1-\frac{\alphat}{2}\Big)\gammat\Big)\frac{\pt}{L}H^t 
	+ \frac{\theta_t^2\pt}{\etat}\Big(D^t -D^{t+1} \Big)	
	\bigg].  \label{eq:key-main}
	\end{align}
\end{lemma}

\subsection{Proof of Theorem~\ref{thm:main}}
\label{sec:proofofthm1}
Now, we are ready to prove the main convergence Theorem~\ref{thm:main}.
According to Lemma~\ref{lem:key-main}, we know the change of the function value after each round. 
By dividing \eqref{eq:key-main} with $\frac{\thetat^2\pt}{\etat}$ on both sides, we obtain
\begin{align}
\EB{\frac{\etat}{\thetat^2\pt}F^{t+1} + \frac{\gammat\etat}{\thetat^2 L}H^{t+1}} \leq \E\bigg[\frac{(1-\thetat\pt)\etat}{\thetat^2\pt}F^t +
\Big(\frac{\omega}{\betat n}+\Big(1-\frac{\alphat}{2}\Big)\gammat\Big)\frac{\etat}{\thetat^2 L}H^t
+ D^t - D^{t+1}	
\bigg]. \label{eq:key-div-main}
\end{align}
Then according to the following conditions on the parameters (see \eqref{eq:parat1} of Theorem~\ref{thm:main}):
\begin{align}
\frac{(1-p_t\theta_t)\eta_t}{p_t\theta_t^2} \leq \frac{\eta_{t-1}}{p_{t-1}\theta_{t-1}^2}, \text{~~and~~}
\Big(\frac{\omega}{\betat n}+\left(1-\frac{\alphat}{2}\right)\gammat\Big)\frac{\etat}{\thetat^2} \leq \frac{\gamma_{t-1}\eta_{t-1}}{\theta_{t-1}^2},~~ \forall t\geq 1. \label{eq:para-cond-main}
\end{align}
The proof of Theorem~\ref{thm:main} is finished by telescoping \eqref{eq:key-div-main} from $t=1$ to $T$ via \eqref{eq:para-cond-main} and maintaining the same inequality \eqref{eq:key-div-main} for $t=0$: 
\begin{align}
\EB{ F^{T+1} + \frac{\gammaT\pT}{ L}H^{T+1}} \leq \frac{\thetaT^2\pT}{\etaT}\bigg(\frac{(1-\theta_0p_0)\eta_0}{\theta_0^2p_0}F^0 +
\Big(\frac{\omega}{\beta_0 n}+\Big(1-\frac{\alpha_0}{2}\Big)\gamma_0\Big)\frac{\eta_0}{\theta_0^2 L}H^0	
+ D^0
\bigg). \label{eq:key-T-main}
\end{align}
\qedb

\newpage
\section{Missing Proof for Theorem~\ref{thm:detail} in Section~\ref{subsec:detailrate}}
\label{app:proofofthm2}

In this appendix, we provide the proof for concrete Theorem~\ref{thm:detail} (which leads to a detailed convergence result). 
First, let us verify that the choice of parameters (i.e., \eqref{eq:set-beta}--\eqref{eq:set-eta}) in Theorem~\ref{thm:detail} satisfies the conditions (i.e., \eqref{eq:two} and \eqref{eq:parat1}) in Theorem~\ref{thm:main}.
According to $p_t$ and $\alpha_t$ in \eqref{eq:set-parameter} and $\gamma_t$ in \eqref{eq:set-beta},
we have
\begin{align}
4\pt\gamma_t \left(1 + \frac{2\pt}{\alphat}\right) = \frac{1}{2}, ~~ \forall t\geq 0. \label{eq:tmp1}  
\end{align}
Then according to \eqref{eq:tmp1}, $\eta_t$ of \eqref{eq:set-eta} and $\alpha_t$ of \eqref{eq:set-parameter}, the first two conditions in \eqref{eq:two} of Theorem~\ref{thm:main} are satisfied, i.e.,
$$\etat \leq \frac{1}{L\left(1+\beta_t+4\pt\gammat\big(1+ \frac{2\pt}{\alphat}\big)\right)} \text{~~and~~}
\alphat\leq \frac{1}{1+\omega},  ~~\forall t\geq 0. $$
Besides, from \eqref{eq:set-beta} and \eqref{eq:set-parameter}, we know that $\theta_t\leq \frac{1}{3}$ and $\frac{2\omega}{\beta_t n}\leq \frac{1}{6}$ for any $t\geq 1$. Combining with \eqref{eq:tmp1}, then the following condition in \eqref{eq:parat1} of Theorem~\ref{thm:main} is satisfied:
$$\frac{2\omega}{\beta_t n}
+ 4\pt\gamma_t \left(1 + \frac{2\pt}{\alphat}\right) \leq 1-\thetat, ~~\forall t\geq1.$$
Now, only the following two conditions in \eqref{eq:parat1} of Theorem~\ref{thm:main} are remained:
\begin{align}
\frac{(1-p_t\theta_t)\eta_t}{p_t\theta_t^2} \leq \frac{\eta_{t-1}}{p_{t-1}\theta_{t-1}^2}, \text{~~and~~}
\Big(\frac{\omega}{\betat n}+\left(1-\frac{\alphat}{2}\right)\gammat\Big)\frac{\etat}{\thetat^2} \leq \frac{\gamma_{t-1}\eta_{t-1}}{\theta_{t-1}^2},~~ \forall t\geq 1. \label{eq:last-two-cond}
\end{align}
For the first condition of \eqref{eq:last-two-cond}, by plugging the parameter choice $\{p_t\}$ and $\{\theta_t\}$ of \eqref{eq:set-parameter}, it is sufficient to let 
\begin{align}
\left(1-\frac{3}{t+9(1+b+\omega)}\right) \etat \leq \left(1-\frac{1}{t+9(1+b+\omega)}\right)^2 \eta_{t-1},~~  \forall t\geq 1. \label{eq:ver1}
\end{align}
For satisfying \eqref{eq:ver1}, it is sufficient to choose $\eta_t$ as in \eqref{eq:set-eta}:
\begin{align}
\eta_t = \min\left\{\Big(1+\frac{1}{t+9(1+b+\omega)}\Big)\eta_{t-1},~ \frac{1}{L(\beta+3/2)}\right\},~~  \forall t\geq 1. \label{eq:eta1}
\end{align}

Similarly, for the second condition of \eqref{eq:last-two-cond}, by plugging the parameter choice $\{\thetat\}$ and $\{\alphat\}$ of \eqref{eq:set-parameter}, it is sufficient to let 
\begin{align}
\Big(\frac{\omega}{\betat n}+\Big(1-\frac{1}{2(1+\omega)}\Big)\gammat\Big)\etat &\leq \gamma_{t-1}\eta_{t-1}\Big(1-\frac{1}{t+9(1+b+\omega)}\Big)^2, ~~  \forall t\geq 1. \label{eq:ver2}
\end{align}
By plugging $\{\beta_t\}$ and $\{\gamma_t\}$ of \eqref{eq:set-beta} into \eqref{eq:ver2}, we have
\begin{align}
\left(1-\frac{1}{3(1+\omega)}\right)\etat &\leq \eta_{t-1}\left(1-\frac{1}{t+9(1+b+\omega)}\right)^2, ~~  \forall t\geq 1. \label{eq:ver3}
\end{align}
Note that the choice of $\etat$ in \eqref{eq:eta1} also satisfies \eqref{eq:ver3}.

Now, we have verified that all conditions of Theorem~\ref{thm:main} are satisfied with the parameter choice in Theorem~\ref{thm:detail}. 
Next, we obtain the detailed convergence results of \canita by using this choice of parameters. 
According to Theorem~\ref{thm:main}, we know that the following equation holds for any $T>0$:
\begin{align}
\EB{F^{T+1} + \frac{\gammaT\pT}{ L}H^{T+1}} \leq \frac{\thetaT^2\pT}{\etaT}\bigg(\frac{(1-\theta_0p_0)\eta_0}{\theta_0^2p_0}F^0 + \Big(\frac{\omega}{\beta_0 n}+ \Big(1-\frac{\alpha_0}{2}\Big)\gamma_0\Big)\frac{\eta_0}{\theta_0^2 L}H^0	
+ D^0\bigg).  \label{eq:T}
\end{align}
According to \eqref{eq:set-parameter}, we have 
\begin{align}
\thetaT^2\pT =\frac{9(1+b)}{(T+9(1+b+\omega))^2}. \label{eq:thetap}
\end{align}
According to \eqref{eq:eta1}, we have 
\begin{align}
\etaT &=  \min\left\{\frac{T+9(1+b+\omega)}{9(1+b+\omega)}\eta_{0},~ \frac{1}{L(\beta+3/2)}\right\} \notag\\
&=\min\left\{\frac{T+9(1+b+\omega)}{9(1+b+\omega)}\frac{1}{L(\beta_0+3/2)},~ \frac{1}{L(\beta+3/2)}\right\} \notag\\
&=\min\left\{\frac{(T+9(1+b+\omega))(1+b)}{162(1+b+\omega)^3},~ \frac{1}{L(\beta+3/2)}\right\}, \label{eq:eta2}
\end{align}
where \eqref{eq:eta2} uses the appropriate $\beta_0 = \frac{9(1+b+\omega)^2}{(1+b)L}$ chosen in \eqref{eq:set-beta} of Theorem~\ref{thm:detail}. 
Besides, according to the initial values of the parameters,
we can simplify the right-hand-side of \eqref{eq:T} with $\frac{(1-\theta_0p_0)\eta_0}{\theta_0^2p_0}\leq 1$ and $\left(1-\frac{\alpha_0}{2}\right)\gamma_0\frac{\eta_0}{\theta_0^2 L}\leq 1$.

Now we plug \eqref{eq:thetap} and \eqref{eq:eta2} into \eqref{eq:T} and omit the constant to obtain
\begin{align}
\EB{F^{T+1}} 
&\leq O\left(\max\left\{\frac{(1+b+\omega)^3}{(T+9(1+b+\omega))^3},~ \frac{(1+b)(\beta +3/2)L}{(T+9(1+b+\omega))^2} \right\}\right)  \notag\\
&\leq O\left(\max\left\{\frac{(1+b+\omega)^3}{T^3},~ \frac{(1+b)(\beta +3/2)L}{T^2} \right\}\right)  \notag\\
&\leq O\left(\max\left\{\frac{(1+\omega)^3}{T^3},~ \frac{(1+\sqrt{\omega(1+\omega)^2/n})L}{T^2} \right\}\right)  \label{eq:plug-b-beta}\\
&=   O\left(\frac{(1+\sqrt{\omega^3/n})L}{T^2} + \frac{\omega^3}{T^3}\right), \label{eq:key-final}
\end{align}
where \eqref{eq:plug-b-beta} uses $b=\min\Big\{\omega, \sqrt{\frac{\omega(1+\omega)^2}{n}}\Big\}$ and $\beta$ of \eqref{eq:set-beta} .
Following from \eqref{eq:key-final}, we know that the number of communication rounds for \canita (Algorithm~\ref{alg:canita}) to find an $\epsilon$-solution such that 
$$ 
\Exp{f(w^{T+1}) - f(x^*)} \overset{\eqref{eq:F,H,D}}{\eqdef}\Exp{F^{T+1}}  \leq \epsilon
$$
is at most 
$$
T= O\left(\sqrt{\bigg(1+\sqrt{\frac{\omega^3}{n}}\bigg)\frac{L}{\epsilon}}
+ \omega\left(\frac{1}{\epsilon}\right)^{\frac{1}{3}} \right).
$$
\qedb

\newpage
\section{Missing Proofs for Six Lemmas in Appendix~\ref{sec:lemmas}}
\label{sec:proofsketch-appendix}

In Appendix~\ref{sec:proofsketch}, we provided the proof of Theorem~\ref{thm:main} using six lemmas. 
Now we present the omitted proofs for these Lemmas~\ref{lem:func-main}--\ref{lem:key-main} in Appendices~\ref{app:proofoflem1}--\ref{app:proofofkeylem}, respectively.

\subsection{Proof of Lemma~\ref{lem:func-main}}
\label{app:proofoflem1}

According to the $L$-smoothness of $f$ (Assumption \ref{asp:smooth}), we have
\begin{align}
&\Exp{f(\bxtn)} \notag\\
&\leq \E\bigg[f(\uxt) + \inner{\nabla f(\uxt)}{\bxtn - \uxt} + \frac{L}{2} \ns{\bxtn-\uxt}\bigg] \notag\\
&= \E\bigg[f(\uxt) + \inner{\nabla f(\uxt)}{\theta_t(\xtn-\xt)} +\frac{L\theta_t^2}{2}\ns{\xtn-\xt}\bigg] \label{eq:use-two-m} \\
&= \E\bigg[f(\uxt) + \inner{\nabla f(\uxt) - \tnabla}{\theta_t(\xtn-\xt)} 
+\inner{\tnabla}{\theta_t(\xtn-\xt)} +\frac{L\theta_t^2}{2}\ns{\xtn-\xt}\bigg]  \notag\\
&\leq \E\bigg[f(\uxt) + \frac{1}{2L\beta_t}\ns{\nabla f(\uxt) - \tnabla} 
+\frac{L\beta_t\theta_t^2}{2}\ns{\xtn-\xt}  
+\frac{L\theta_t^2}{2}\ns{\xtn-\xt}  
\notag\\
&\qquad \qquad \qquad 
+\inner{\tnabla}{\theta_t(\xtn-\xt)} \bigg]   \label{eq:useyoung} \\
&= \E\bigg[f(\uxt) + \frac{1}{2L\beta_t}\ns{\nabla f(\uxt) - \tnabla} 
+\frac{L(1+\beta_t)\theta_t^2}{2}\ns{\xtn-\xt} 
\notag\\
&\qquad \qquad \qquad 
+\inner{\tnabla}{\theta_t(x^*-\xt)}
+\inner{\tnabla}{\theta_t(\xtn-x^*)} \bigg]  \notag\\
&= \E\bigg[f(\uxt) + \frac{1}{2L\beta_t}\ns{\nabla f(\uxt) - \tnabla} 
+\frac{L(1+\beta_t)\theta_t^2}{2}\ns{\xtn-\xt} 
+\inner{\nabla f(\uxt)}{\theta_t(x^*-\xt)}
\notag\\
&\qquad \qquad \qquad 
+\inner{\tnabla}{\theta_t(\xtn-x^*)} \bigg] \label{eq:use-exp}\\
&=\E\bigg[f(\uxt) + \frac{1}{2L\beta_t}\ns{\nabla f(\uxt) - \tnabla} 
+\frac{L(1+\beta_t)\theta_t^2}{2}\ns{\xtn-\xt} 
+\inner{\nabla f(\uxt)}{\theta_t(x^*-\xt)}
\notag\\
&\qquad \qquad \qquad 
+\frac{\theta_t^2}{\etat}\inner{\xt-\xtn}{\xtn-x^*} \bigg] \label{eq:use-update}\\
&=\E\bigg[f(\uxt) + \frac{1}{2L\beta_t}\ns{\nabla f(\uxt) - \tnabla} 
+\frac{L(1+\beta_t)\theta_t^2}{2}\ns{\xtn-\xt} 
+\inner{\nabla f(\uxt)}{\theta_t(x^*-\xt)}
\notag\\
&\qquad \qquad \qquad 
+\frac{\theta_t^2}{2\etat}\big(\ns{\xt-x^*}-\ns{\xt-\xtn}-\ns{\xtn-x^*} \big) \bigg] \notag\\
&=\E\bigg[f(\uxt) +\inner{\nabla f(\uxt)}{\theta_t(x^*-\xt)} +\frac{\theta_t^2}{2\etat}\big(\ns{\xt-x^*}-\ns{\xtn-x^*} \big)	 \notag\\
&\qquad \qquad \qquad 
-\Big(\frac{\theta_t^2}{2\etat}-\frac{L(1+\beta_t)\theta_t^2}{2}\Big)\ns{\xtn-\xt} 
+\frac{1}{2L\beta_t}\ns{\nabla f(\uxt) - \tnabla} 
\bigg], \notag
\end{align}
where \eqref{eq:use-two-m} holds since $\bxtn - \uxt = \thetat (\xtn - \xt)$ according to the two momentum/interpolation steps of \canita (see Line~\ref{line:uxt} and Line~\ref{line:bxtn} of Algorithm~\ref{alg:canita}), \eqref{eq:useyoung} uses Young's inequality with any $\beta_t>0$, \eqref{eq:use-exp} holds due to $\E[\tnabla]=\nabla f(\uxt)$ since the compression is unbiased from \eqref{eq:comp}, and \eqref{eq:use-update} holds according to the gradient update step $\xtn = \xt - \frac{\etat}{\thetat} \gt$ (see Line~\ref{line:update} of Algorithm~\ref{alg:canita}).
\qedb

\subsection{Proof of Lemma~\ref{lem:var-main}}

This lemma is proved as follows:
\begin{align}
\Exp{\ns{\nabla f(\uxt) -\tnabla}}
&=
\E\left[\nsB{\frac{1}{n}\sum_{i=1}^n\Big(\cit(\nabla f_i(\uxt) -\hit) + \hit - \nabla f_i(\uxt)\Big)}\right] \notag\\
&=
\frac{1}{n^2}\sum_{i=1}^n \E\left[\nsB{\cit(\nabla f_i(\uxt) - \hit) + \hit - \nabla f_i(\uxt)}\right] \notag\\
&\leq
\frac{\omega}{n^2}\sum_{i=1}^n \ns{\nabla f_i(\uxt) - \hit} \label{eq:use-comp} \\
&\leq
\frac{2\omega}{n^2}\sum_{i=1}^n \ns{\nabla f_i(\uxt) - \nabla f_i(\wt)}	
+ \frac{2\omega}{n^2}\cht, \label{eq:use-cauchy}
\end{align}
where \eqref{eq:use-comp} follows from the definition of $\omega$-compression operator (i.e., \eqref{eq:comp}), and the last inequality \eqref{eq:use-cauchy} uses Cauchy-Schwarz inequality.
\qedb

\subsection{Proof of Lemma~\ref{lem:shift-main}}

Firstly, according to the probabilistic update of $\wtn$ (see Line~\ref{line:prob} of Algorithm~\ref{alg:canita}) and recalling that $H^t \eqdef \frac{1}{n}\cht$ defined in \eqref{eq:F,H,D}, we get
\begin{align}
&\EB{H^{t+1}} \notag\\
&=
\frac{\pt}{n}\sum_{i=1}^n\EB{\norm{\nabla f_i(\bxtn)-\hitn}^2}
+
\frac{1-\pt}{n}\sum_{i=1}^n\EB{\norm{\nabla f_i(\wt)-\hitn}^2} \notag\\
&\leq
\left(1 + \frac{2\pt}{\alphat}\right)\frac{\pt}{n}\sum_{i=1}^n\EB{\norm{\nabla f_i(\bxtn) - \nabla f_i(\wt)}^2}
+\left(1 + \frac{\alphat}{2\pt}\right)\frac{\pt}{n}\sum_{i=1}^n\EB{\norm{ \nabla f_i(\wt)-\hitn}^2}  \notag\\
&\qquad \qquad \qquad 
+\frac{1-\pt}{n} \sum_{i=1}^n\EB{\norm{ \nabla f_i(\wt)-\hitn}^2} .
\label{eq:usecauchy1}\\ 
&\leq
\left(1 + \frac{2\pt}{\alphat}\right)\frac{\pt}{n}\sum_{i=1}^n\EB{\norm{\nabla f_i(\bxtn) - \nabla f_i(\wt)}^2}
+ \Big(1+\frac{\alphat}{2}\Big)\Big(1-2\alphat+\alphat^2(1+\omega)\Big) H^t  \label{eq:useh}\\
&\leq
\left(1 + \frac{2\pt}{\alphat}\right)\frac{\pt}{n}\sum_{i=1}^n\EB{\norm{\nabla f_i(\bxtn) - \nabla f_i(\wt)}^2}
+ \left(1-\frac{\alphat}{2}\right) H^t\label{eq:usealpha}\\
&\leq \left(1 + \frac{2\pt}{\alphat} \right)\frac{2\pt}{n}\sum_{i=1}^n\EB{\norm{\nabla f_i(\bxtn) - \nabla f_i(\uxt)}^2 
	+ \norm{\nabla f_i(\uxt) - \nabla f_i(\wt)}^2} 
+ \left(1-\frac{\alphat}{2}\right) H^t \label{eq:usecauchy2}\\
&\leq \left(1 + \frac{2\pt}{\alphat} \right)\frac{2\pt}{n}\sum_{i=1}^n\EB{L^2\norm{ \bxtn -\uxt}^2 
	+ \norm{\nabla f_i(\uxt) - \nabla f_i(\wt)}^2} 
+ \left(1-\frac{\alphat}{2}\right) H^t  \label{eq:usesmooth}\\
&\leq 2\pt L^2\thetat^2\left(1 + \frac{2\pt}{\alphat} \right)\Exp{\ns{\xtn -\xt}}
+ 2\pt\left(1 + \frac{2\pt}{\alphat} \right)Y^t 
+ \left(1-\frac{\alphat}{2}\right)H^t, \label{eq:last}
\end{align}
where \eqref{eq:usecauchy1} uses Young's inequality, 
\eqref{eq:useh} uses the update of local shifts $\hitn =\hit+\alphat \cit(\nabla f_i(\wt) - \hit)$ (see Line~\ref{line:shifth} of Algorithm~\ref{alg:canita}) and the property of $\omega$-compression operator (i.e., \eqref{eq:comp}), \eqref{eq:usealpha} uses $\alphat\leq \nicefrac{1}{(1+\omega)}$, \eqref{eq:usecauchy2} uses Cauchy-Schwarz inequality, \eqref{eq:usesmooth} uses the $L$-smoothness of $f_i$ (Assumption \ref{asp:smooth}), and the last inequality \eqref{eq:last} holds since $\bxtn - \uxt = \thetat (\xtn - \xt)$ according to the two interpolation steps of \canita (see Line~\ref{line:uxt} and Line~\ref{line:bxtn} of Algorithm~\ref{alg:canita}).
\qedb

\subsection{Proof of Lemma~\ref{lem:gradfunc-main}}

This lemma directly follows from a standard result under Assumption~\ref{asp:smooth}. 
According to e.g.  Lemma~1 of \citep{Zhize2019unified} or Lemma~5 of \citep{li2021anita}, we have 
\begin{align}
\frac{1}{2L} \|\nabla f_i(\wt) - \nabla f_i(\uxt)\|^2 \leq
f_i(\wt) - f_i(\uxt) - \langle \nabla f_i(\uxt), \wt - \uxt \rangle. \label{eq:l-anita}
\end{align}
Then, the result \eqref{eq:gradfunc-main} is obtained by summing up \eqref{eq:l-anita} for all $i\in [n]$ and noting $f(x):=\frac{1}{n}\sum_{i=1}^{n}f_i(x)$ (see \eqref{eq:prob}) and $Y^t \eqdef \frac{1}{n}\sum_{i=1}^n \ns{\nabla f_i(\wt)-\nabla f_i(\uxt)}$ (see \eqref{eq:defYt}).
\qedb

\subsection{Proof of Lemma~\ref{lem:fwfx-main}}

The lemma follows directly from the probabilistic update of $\wtn$; see Line~\ref{line:prob} of Algorithm~\ref{alg:canita}.
\qedb

\subsection{Proof of Lemma~\ref{lem:key-main}}
\label{app:proofofkeylem}
Now, we provide the detailed proof for the key Lemma~\ref{lem:key-main} by using previous Lemmas~\ref{lem:func-main}--\ref{lem:fwfx-main}.
First, we plug \eqref{eq:var-main} of Lemma~\ref{lem:var-main} into \eqref{eq:func-main} of Lemma~\ref{lem:func-main} to obtain
\begin{align}
\Exp{f(\bxtn)} &\leq \E\bigg[f(\uxt) +\inner{\nabla f(\uxt)}{\theta_t(x^*-\xt)} +\frac{\theta_t^2}{\etat}\big(D^t - D^{t+1} \big)	 \notag\\
&\qquad 
-\left(\frac{\theta_t^2}{2\etat}-\frac{L(1+\beta_t)\theta_t^2}{2}\right)\ns{\xtn-\xt} 
+\frac{\omega}{L\beta_t n} Y^t	+ \frac{\omega}{L\beta_tn}H^t \bigg]. \label{eq:key1-main}
\end{align}
Then, we add \eqref{eq:key1-main} and $\frac{\gammat}{L} \times$ \eqref{eq:shift-main} of Lemma~\ref{lem:shift-main} to get
\begin{align}
&\EB{f(\bxtn) + \frac{\gammat}{L}H^{t+1}}  \notag\\
&\leq \E\bigg[f(\uxt) +\inner{\nabla f(\uxt)}{\theta_t(x^*-\xt)} +\frac{\theta_t^2}{\etat}\big(D^t - D^{t+1} \big)	 \notag\\
&\qquad \qquad
-\left(\frac{\theta_t^2}{2\etat}-\frac{L(1+\beta_t)\theta_t^2}{2}\right)\ns{\xtn-\xt}  
+\frac{\omega}{L\beta_t n}Y^t	
+\frac{\omega}{L\betat n} H^t \notag\\
&\qquad \qquad
+\left(1-\frac{\alphat}{2}\right)\frac{\gammat}{L}H^t 
+ \left(1 + \frac{2\pt}{\alphat} \right)\frac{2\pt\gammat}{L}Y^t 
+2\pt \gammat L\thetat^2\left(1 + \frac{2\pt}{\alphat} \right)\ns{\xtn -\xt}
\bigg] \notag\\
&=  \E\bigg[f(\uxt) 	  
+\inner{\nabla f(\uxt)}{\theta_t(x^*-\xt)} +\frac{\theta_t^2}{\etat}\big(D^t - D^{t+1} \big)	 \notag\\
&\qquad \qquad
-\left(\frac{\theta_t^2}{2\etat}-\frac{L(1+\beta_t)\theta_t^2}{2}
- 2\pt \gammat L\thetat^2\Big(1 + \frac{2\pt}{\alphat} \Big)\right)\ns{\xtn-\xt}  \notag\\
&\qquad \qquad
+\left(\frac{\omega}{\betat n}+\left(1-\frac{\alphat}{2}\right)\gammat\right)\frac{1}{L}H^t 
+\left(\frac{2\omega}{\beta_t n}
+ 4\pt\gamma_t\Big(1 + \frac{2\pt}{\alphat}\Big) \right)
\frac{1}{2L}Y^t
\bigg]  \notag\\
&\leq  \E\bigg[f(\uxt) 	  
+\inner{\nabla f(\uxt)}{\theta_t(x^*-\xt)} +\frac{\theta_t^2}{\etat}\left(D^t - D^{t+1} \right)	 
+\left(\frac{\omega}{\betat n}+\left(1-\frac{\alphat}{2}\right)\gammat\right)\frac{1}{L}H^t \notag\\
&\qquad \qquad
+\left(\frac{2\omega}{\beta_t n}
+ 4\pt\gamma_t\Big(1 + \frac{2\pt}{\alphat}\Big) \right)
\frac{1}{2L} Y^t 
\bigg]  \label{eq:use-etat-main} \\
&\leq  \E\bigg[f(\uxt) 	  
+\inner{\nabla f(\uxt)}{\theta_t(x^*-\xt)} +\frac{\theta_t^2}{\etat}\big(D^t - D^{t+1}\big)	 
+\left(\frac{\omega}{\betat n}+\left(1-\frac{\alphat}{2}\right)\gammat\right)\frac{1}{L}H^t \notag\\
&\qquad \qquad
+ \frac{1-\thetat}{2L} Y^t
\bigg]  \label{eq:use-thetat-main} \\
&\leq  \E\bigg[f(\uxt) 	  
+\inner{\nabla f(\uxt)}{\theta_t(x^*-\xt)} +\frac{\theta_t^2}{\etat}\big(D^t - D^{t+1} \big)	 
+\left(\frac{\omega}{\betat n}+\left(1-\frac{\alphat}{2}\right)\gammat\right)\frac{1}{L}H^t \notag\\
&\qquad \qquad
+ (1-\thetat)\Big(f(\wt) - f(\uxt) - \inner{\nabla f(\uxt)}{\wt-\uxt}\Big)
\bigg]  \label{eq:use-gradfunc-main} \\
&=  \E\bigg[f(\uxt) 	  
+\inner{\nabla f(\uxt)}{\theta_t(x^*-\xt)} +\frac{\theta_t^2}{\etat}\big(D^t - D^{t+1} \big)	 
+\left(\frac{\omega}{\betat n}+\left(1-\frac{\alphat}{2}\right)\gammat\right)\frac{1}{L}H^t \notag\\
&\qquad \qquad
+ (1-\thetat)\left(f(\wt) - f(\uxt)\right) - \thetat\inner{\nabla f(\uxt)}{\uxt-\xt} 
\bigg]  \label{eq:use-wtuxt-main} \\
&\leq  \E\bigg[  (1-\thetat)f(\wt) +\thetat f(x^*)  
+\frac{\theta_t^2}{\etat}\left(D^t-D^{t+1} \right)	+\left(\frac{\omega}{\betat n}+\left(1-\frac{\alphat}{2}\right)\gammat\right)\frac{1}{L}H^t 
\bigg],  \label{eq:use-convex-main} 
\end{align}
where \eqref{eq:use-etat-main} holds by letting $\etat \leq \frac{1}{L\big(1+\beta_t+4\pt\gammat(1+2\pt/\alphat)\big)}$, 
\eqref{eq:use-thetat-main} holds by letting  $\frac{2\omega}{\beta_t n}
+ 4\pt\gamma_t(1 + \frac{2\pt}{\alphat}) \leq 1-\thetat$,
\eqref{eq:use-gradfunc-main} follows from \eqref{eq:gradfunc-main} of Lemma~\ref{lem:gradfunc-main},
\eqref{eq:use-wtuxt-main} holds since $\uxt = \thetat \xt + (1 - \thetat)\wt$ (see Line~\ref{line:uxt} of Algorithm~\ref{alg:canita}),
and the last inequality \eqref{eq:use-convex-main} uses the convexity of $f$.
Also note that \eqref{eq:use-thetat-main} from \eqref{eq:use-etat-main} uses $\frac{2\omega}{\beta_t n}
+ 4\pt\gamma_t(1 + \frac{2\pt}{\alphat}) \leq 1-\thetat$, however this condition is only needed for $t\geq 1$, i.e., it is not needed for the case $t=0$ since $Y^0=0$ from $\ux^0=w^0=x^0$. The function and inner product terms will also perform the same result in the final \eqref{eq:use-convex-main} since $\ux^0=w^0=x^0$.

The proof of Lemma~\ref{lem:key-main} is finished by adding $\eqref{eq:use-convex-main} \times \pt$ and \eqref{eq:fwfx-main} of Lemma~\ref{lem:fwfx-main} to obtain \eqref{eq:key-main}.	
\qedb

\end{document}